\pdfoutput=1
\documentclass[11pt]{article}

\usepackage[final]{acl}

\usepackage{times}
\usepackage{latexsym}
\usepackage[utf8]{inputenc} % allow utf-8 input
\usepackage[T1]{fontenc}    % use 8-bit T1 fonts
\usepackage{hyperref}       % hyperlinks
\usepackage{url}            % simple URL typesetting
\usepackage{booktabs}       % professional-quality tables
\usepackage{amsfonts}       % blackboard math symbols
\usepackage{nicefrac}       % compact symbols for 1/2, etc.
\usepackage{microtype}      % microtypography
\usepackage{xcolor}         % colors
\usepackage{amsmath}
\usepackage{amsfonts}
\usepackage{multirow}
\usepackage{subfigure}
\usepackage{subcaption}
\usepackage{pgfplots}
\usepackage{xspace}
\usepackage{enumitem}
\usepackage{wrapfig}
\usepackage{bbding}
\usepackage{colortbl}
\usepackage{adjustbox}
\usepackage[T1]{fontenc}
\usepackage[utf8]{inputenc}

\usepackage{microtype}

\usepackage{inconsolata}

\usepackage{graphicx}

\title{MPO: Multilingual Safety Alignment via Reward Gap Optimization}

\author{Weixiang Zhao$^1$, Yulin Hu$^1$, Yang Deng$^2$, Tongtong Wu$^3$, Wenxuan Zhang$^4$ \\ \textbf{Jiahe Guo}$^1$, \textbf{An Zhang}$^5$, \textbf{Yanyan Zhao}$^1$\thanks{\ \ Corresponding author}, \textbf{Bing Qin}$^1$, \textbf{Tat-Seng Chua}$^5$, \textbf{Ting Liu}$^1$ \\
        $^1$Harbin Institute of Technology, $^2$Singapore Management University,
        $^3$Monash University \\ $^4$Singapore University of Technology and Design,
        $^5$National University of Singapore\\
        \texttt{\{wxzhao, yyzhao\}@ir.hit.edu.cn}}

\begin{document}
\maketitle
\begin{abstract}

Large language models (LLMs) have become increasingly central to AI applications worldwide, necessitating robust multilingual safety alignment to ensure secure deployment across diverse linguistic contexts. Existing preference learning methods for safety alignment, such as RLHF and DPO, are primarily monolingual and struggle with noisy multilingual data. To address these limitations, we introduce \textbf{\underline{M}}ultilingual reward ga\textbf{\underline{P}} \textbf{\underline{O}}ptimization (\textbf{MPO}), a novel approach that leverages the well-aligned safety capabilities of the dominant language (\textit{e.g.}, English) to improve safety alignment across multiple languages. MPO directly minimizes the reward gap difference between the dominant language and target languages, effectively transferring safety capabilities while preserving the original strengths of the dominant language. Extensive experiments on three LLMs, LLaMA-3.1, Gemma-2 and Qwen2.5, validate MPO’s efficacy in multilingual safety alignment without degrading general multilingual utility. Our code is available at: \url{https://github.com/circle-hit/MPO}. \textcolor{red}{WARNING: This paper may contain content that is offensive and harmful.}

\end{abstract}

\section{Introduction}

Large language models (LLMs) are increasingly driving global applications \citep{brown2020language,touvron2023llama1,touvron2023llama2,jiang2023mistral,dubey2024llama,team2024gemma}, enabling users from diverse linguistic and cultural backgrounds to access the benefits of AI advancements \citep{zhaowildchat,zhenglmsys}. In this context, achieving multilingual safety alignment is crucial to ensuring secure deployment across various languages \citep{kanepajs2024towards,friedrich2024llms}. However, recent studies highlight substantial differences in the safety challenges faced by LLMs across various languages, with models being more prone to generate unsafe responses in low-resource languages. \citep{yong2023low,deng2024multilingual,wang-etal-2024-languages,shen2024language}.

To mitigate such challenge, one straightforward solution is to conduct safety preference alignment for each language, with methods like reinforcement learning from human feedback (RLHF) \citep{ouyang2022training} or direct preference optimization (DPO) \citep{rafailov2023direct}.

However, a key issue is the scarcity of multilingual data available \citep{ahmadian2024multilingual,wu2024reuse,hong2024cross}. Though off-the-shelf translation tools could be employed to generate training data in various languages, the resulting translations---especially for low-resource languages---are often noisy, riddled with unusual phrasing and inaccurate content \citep{zhang2024enhancing,liu2024translation}. On the other hand, current prevailing preference learning paradigms are highly sensitive to noisy data \citep{bai2022training,wang2024secrets,chowdhury2024provably,alfano2024learning}. In some cases, such noise-induced errors may even cause safety misalignment \citep{shen2024language,razin2024unintentional}, further exacerbating multilingual safety concerns.

To address this challenge, we first conduct an empirical analysis on several widely-used LLMs, including LLaMA-3.1 \citep{dubey2024llama}, Gemma-2 \citep{team2024gemma}, and Qwen2.5 \citep{yang2024qwen2}, which have undergone sufficient safety alignment for their dominant language (typically English). We identify a crucial pattern: the implicit reward gap---defined as the log-likelihood difference between safe and unsafe responses---strongly correlates with multilingual safety performance. The dominant language (English) exhibits a substantially larger reward gap (RG) compared to low-resource ones, directly corresponding to its superior safety performance measured by Attack Success Rate (ASR). This inverse RG-ASR relationship establishes the reward gap as a quantifiable indicator of safety alignment quality across languages.

\begin{figure}[t]
\centering
\includegraphics[width=1.0\columnwidth]{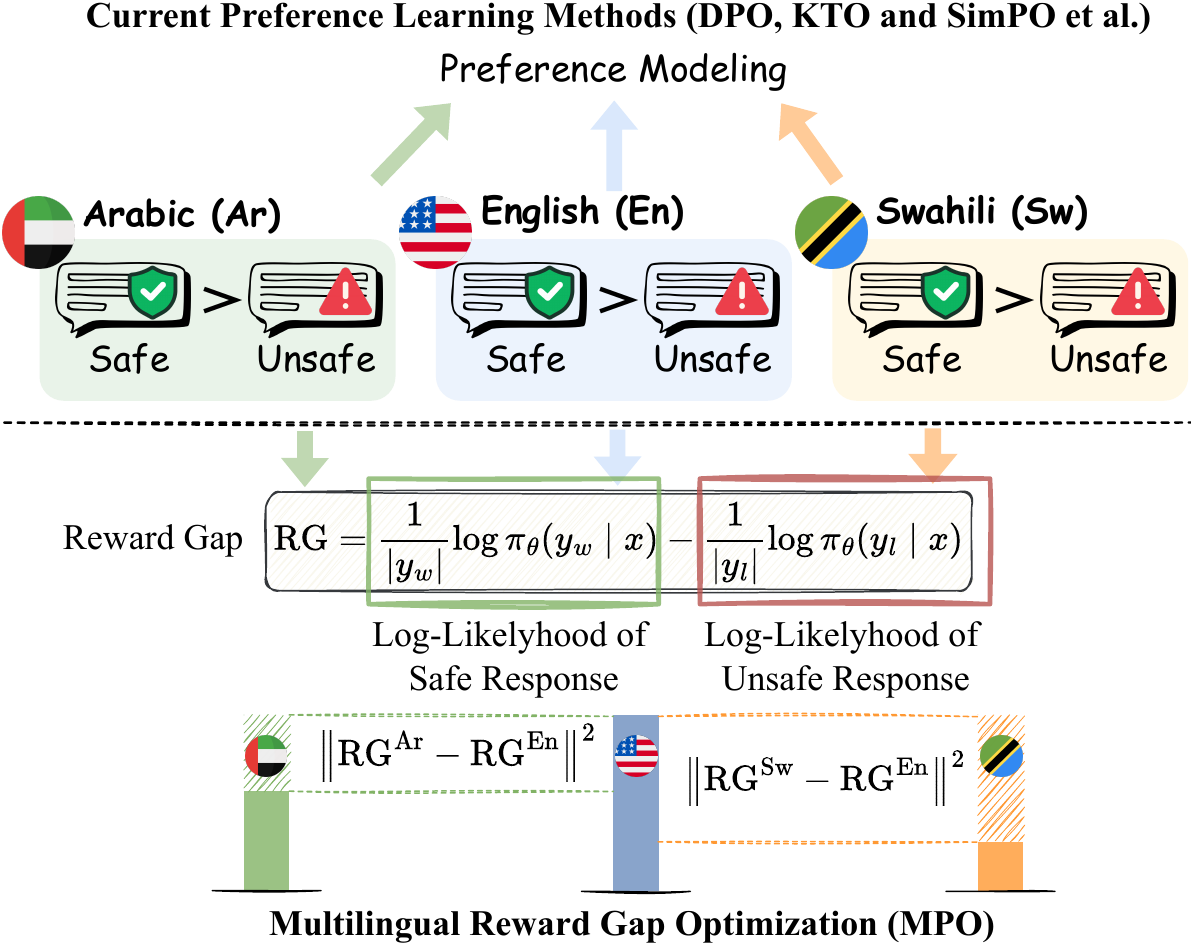}
\caption{Top: Current preference learning methods optimize noisy multilingual preference data. Bottom: Our MPO directly minimizes the discrepancy of reward gap across different languages.}
\label{fig:motivation}
\end{figure}

Building on these insights, we propose \underline{\textbf{M}}ultilingual reward ga\underline{\textbf{P}} \underline{\textbf{O}}ptimization (\textbf{MPO}), a novel alignment paradigm for multilingual safety challenge that transfers safety capabilities from well-aligned dominant languages to others through reward gap optimization.
As shown in Figure \ref{fig:motivation}, unlike conventional preference learning approaches that attempt to directly optimize noisy multilingual preference data, MPO instead minimizes the discrepancy between the dominant language's robust and well-established reward gap and target languages' weaker alignment signals. To preserve the capabilities of the dominant language from degradation, we also incorporate constraints that maintain its hidden representations largely intact.

Our extensive experiments on LLaMA-3.1-8B-Instruct, Gemma-2-9B-it and Qwen2.5-7B-Instruct, showcase the efficacy and scalability of MPO in multilingual safety alignment over current preference learning methods without compromising the general multilingual utility. Deeper analysis reveals that MPO consistently outperforms across training datasets of varying quality. This further confirms that the reward gap of the dominant language serves as a more reliable and scalable supervision signal for effective multilingual safety alignment.

The main contributions of this work are summarized as follows: 
\begin{itemize}[leftmargin=*]
    \item We propose to leverage the well-aligned safety capabilities of the dominant language as a high-quality supervision signal for multilingual safety alignment.
    \item We propose MPO, which directly minimizes the reward gap difference between the dominant language and target languages, enabling effective multilingual safety alignment.
    \item Experiments on three backbones demonstrate the superior performance of MPO over existing preference learning methods.
\end{itemize}

\section{Preliminaries}

In this section, we first introduce the formulation for the implicit reward gap re-parameterized by DPO  (\S\ref{subsec:dpo}), as well as its improvements and optimizations in SimPO (\S\ref{subsec:simpo}). Specifically, we offer their corresponding interpretations in the context of \textbf{\textit{multilingual safety alignment}}.

\subsection{Direct Preference Optimization (DPO)}
\label{subsec:dpo}
DPO \citep{rafailov2023direct} is one of the most widely used methods for preference learning in LLM alignment. Unlike approaches that involve training an explicit reward model \citep{ouyang2022training}, DPO re-parameterizes the implicit reward function $r$ using a closed-form expression derived from the Bradley-Terry (BT) model \citep{bradley1952rank} with the optimal policy:  
\begin{equation}\small
\label{eq:dpo_reward}
r(x,y) = \beta \log \frac{\pi_\theta(y \mid x)}{\pi_{\text{ref}}(y \mid x)} + \beta \log Z(x),
x\end{equation}
where $\pi_\theta$ is the policy model, $\pi_{\text{ref}}$ is the reference model, typically the supervised fine-tuned (SFT) checkpoint, $\beta$ is a hyper-parameter and $Z(x)$ is the partition function.

In the context of multilingual safety alignment, the reward gap of the backbone model between safe and unsafe responses in different languages $t$ can be expressed as:
\begin{equation}\small
\begin{split}
    \text{RG}^t &=  r(x^t,y_w^t) -  r(x^t,y_l^t) \\
    &=\beta \log \frac{\pi_{\theta}(y_w^t|x^t)}{\pi_{\text{ref}} (y_w^t|x^t)} -  \beta \log \frac{\pi_{\theta}(y_l^t|x^t)}{\pi_{\text{ref}} (y_l^t|x^t)} \label{eq:dpo_rg}
\end{split}
\end{equation}
where the triplet $(x^t, y_w^t, y_l^t)$ are preference pairs related to safety concerns in language $t$, consisting of the input query $x^t$, the preferred (safe) response $y_w^t$, and the dispreferred (unsafe) response $y_l^t$.

\subsection{Simple Preference Optimization (SimPO)}
\label{subsec:simpo}

As pointed out by SimPO \citep{meng2024simpo}, using Eq. (\ref{eq:dpo_reward}) as the implicit reward has the following drawbacks: it creates a mismatch between the reward optimized in training and the log-likelihood optimized during inference. To address this issue, SimPO considers using the average log-likelihood as the implicit reward:
\begin{equation}\small
\label{eq:avg_likelihood}
 r(x,y) = p_\theta(y \mid x) = \frac{1}{|y|} \log \pi_\theta(y \mid x)
\end{equation}

Accordingly, the reward gap is formulated as:
\begin{equation}\small
\label{eq:simpo_rg}
    \text{RG}^t =  \frac{1}{|y_w^t|} \log \pi_\theta(y_w^t \mid x^t) -  \frac{1}{|y_l^t|} \log \pi_\theta(y_l^t \mid x^t) 
\end{equation}

We posit that, compared with Eq. (\ref{eq:dpo_rg}), the reward gap in Eq. (\ref{eq:simpo_rg}) provides a more accurate measure of safety performance differences across languages due to the following reasons: (1) It aligns with the likelihood metric that governs response generation, where a larger reward gap signifies a higher probability of producing safe responses over unsafe ones, serving as a direct indicator of safety performance. (2) Length normalization mitigates reward errors caused by length bias \citep{singhal2023long,park2024disentangling}---unsafe responses, which frequently include specific harmful content, are often longer than safe responses, which typically exhibit concise refusal patterns. Please refer to Appendix \ref{app:rg_discuss} for more emperical evidence and discussion.

\begin{table}
\centering
\resizebox{\linewidth}{!}{
\begin{tabular}{cccccccc}
\toprule 
& & \textbf{En} & \textbf{Zh} & \textbf{Ko} & \textbf{Ar} & \textbf{Bn} & \textbf{Sw} \\
\midrule
\multirow{2}{*}{\textbf{LLaMA-3.1}} &\textbf{RG}$\uparrow$ & \textbf{1.58} &0.36 &0.29 &0.60 &0.04 &0.05 \\
&\textbf{ASR}$\downarrow$ &\textbf{9.00}	&22.00	&50.00	&15.00	&55.00	&57.00 \\
\midrule
\multirow{2}{*}{\textbf{Gemma-2}} &\textbf{RG}$\uparrow$ &\textbf{2.32} &0.69 &0.44 &0.76 &0.42 &0.41 \\
&\textbf{ASR}$\downarrow$ &\textbf{0.00}	&9.00	&14.00	&4.00	&24.00	&26.00 \\
\midrule
\multirow{2}{*}{\textbf{Qwen-2.5}} &\textbf{RG}$\uparrow$ &\textbf{1.87} &\textbf{1.81} &0.69 &0.78 &0.14 &0.20 \\
&\textbf{ASR}$\downarrow$ &\textbf{13.00} &\textbf{9.00}	&21.00	&20.00	&69.00	&98.00 \\
\bottomrule 
\end{tabular}
}
\caption{Results of reward gap (RG) and safety performance across six languages. The evaluation metric used for safety is the Attack Success Rate (ASR), where lower values indicate better performance. Results of the dominant languages are highlighted in bold.}
\label{tab:reward_gap}
\end{table}

\section{Multilingual Reward Gap Optimization}
In this section, we first demonstrate the relationship between the reward gap and the multilingual safety performance for different languages on three backbone LLMs (\S\ref{subsec:reward_gap}). Then, we derive the MPO objective (\S\ref{subsec:MPO_obj}) and perform gradient analysis (\S\ref{subsec:gradient}).

\subsection{Reward Gap across Languages}
\label{subsec:reward_gap}
\paragraph{Models} We select two English-centric LLMs: LLaMA-3.1-8B-Instruct \citep{dubey2024llama} and Gemma-2-9B-it \citep{team2024gemma} and one bilingual LLMs: Qwen2.5-7B-Instruct \citep{yang2024qwen2}, to demonstrate the reward gap (Eq. (\ref{eq:simpo_rg})) on safety issues across different languages.

\paragraph{Languages} We select six languages for evaluation based on the availability of language resources. The high-resource languages are English (En) and Chinese (Zh); the medium-resource languages are Korean (Ko) and Arabic (Ar); and the low-resource languages are Bengali (Bn) and Swahili (Sw). For LLaMA-3.1-8B-Instruct and Gemma-2-9B-it, En serves as the dominant language, while that for Qwen2.5-7B-Instruct is En and Zh.

\paragraph{Data} We utilize the PKU-SafeRLHF dataset \citep{ji2024pku} for the reward gap evaluation across languages. This dataset comprises high-quality English preference pairs focused on safety-related questions. To extend its scope, we randomly sample 100 instances and translate them into each target language using the Google Translate API. Subsequently, we query LLMs directly with these multilingual inputs. The reward gap is computed using Eq. (\ref{eq:simpo_rg}), while safety performance is evaluated based on the Attack Success Rate (ASR).

\paragraph{Analysis} According to the results in Table \ref{tab:reward_gap}, we can draw two key insights:
\begin{itemize}[leftmargin=*]
    \item \textbf{\textit{Inverse relationship between RG and ASR}}: Higher RG corresponds to lower ASR, indicating better safety performance. This demonstrates that RG can, to some extent, reflect the safety performance of LLMs in a specific language.
    \item \textit{\textbf{Safety performance varies significantly across languages}}: As reflected in RG values, lower-resource languages exhibit significantly lower RG compared to high-resource dominant ones, underscoring critical safety concerns in lower-resource settings.
\end{itemize}

\subsection{The MPO Objective}
\label{subsec:MPO_obj}

Based on the above insights, we propose a novel method for multilingual safety alignment called \textbf{\underline{M}}ultilingual reward ga\textbf{\underline{P}} \textbf{\underline{O}}ptimization (\textbf{MPO}). It takes the high-quality and well-aligned RG of the dominant language in LLMs as the pivot and aligns the RG of the target language to it. This facilitates the transfer of the dominant language’s safety capabilities to the target language. This process can be formulated as:
\begin{equation}\small
\label{MPO_loss}
    \mathcal{L}_1 = \mathbb{E}_{(x, y_w, y_l) \sim \mathcal{D}} \left[ \left\Vert    
    \beta \, \text{RG}^t - \text{RG}^d
    \right\Vert^2 \right]
\end{equation}
where $t$ and $d$ represent target and dominant languages, respectively. $\beta$ functions to balance and stabilize the optimization \citep{haarnoja2018soft}.

And RG is calculated by:
\begin{equation}\small
\label{source_gap_t}
    \text{RG}^t = \frac{1}{|y^t_w|} \log\pi_\theta(y^t_w|x^t) - \frac{1}{|y^t_l|} \log\pi_\theta(y^t_l|x^t)
\end{equation}
\begin{equation}\small
\label{source_gap_d}
    \text{RG}^d = \frac{1}{|y^d_w|} \log\pi_{\text{ref}}(y^d_w|x^d) - \frac{1}{|y^d_l|} \log\pi_{\text{ref}}(y^d_l|x^d)
\end{equation}
where the triplets $(x^t, y_w^t, y_l^t)$ and $(x^d, y_w^d, y_l^d)$ are preference pairs derived from target and dominant languages, respectively. Here, $\pi_\theta$ denotes the policy model, while $\pi_{\text{ref}}$ serves as the reference model.

To ensure that the capabilities of the dominant language are not compromised, we constrain the representations of dominant language (at the position of the last token) to remain largely intact:
\begin{gather}\small
\label{eq:rep_cons}
    \mathcal{L}_2 = \mathbb{E}_{x^d \sim \mathcal{D}} \left[ \left\|\boldsymbol{h}^d - \boldsymbol{h}^d_{\text{ref}}\right\|^{2}\right]
\end{gather}
where $\boldsymbol{h}^d_{\text{ref}}$ is the representation of dominant language $x^d$ obtained from the reference model. Inspired by recent empirical findings suggesting that modifying the hidden representations of LLMs is more effective for behavior control \citep{zou2023representation}, we choose to constrain these representations directly, rather than applying KL-based regularization on logits \citep{ziegler2019fine}.

The final optimization objective of MPO is:
\begin{gather}\small
    \mathcal{L} = \mathcal{L}_1 + \mathcal{L}_2
\end{gather}

\subsection{What does the MPO update do?}
\label{subsec:gradient}
The gradient for the learning of target languages with respect to the parameter $\theta$ can be written as:
\begin{equation}\small
\nabla_\theta \,\mathcal{L}_1(\theta)=
2\beta\,\mathbb{E}_{(x, y_w, y_l)\,\sim\,\mathcal{D}}
\biggl(
     w_{\theta} \, \nabla_\theta \,\text{RG}^{t}(\theta)
\biggr)
\end{equation}
where $\nabla_\theta \,\text{RG}^{t}(\theta)$ increases the likelihood of the preferred (safe) response $y_w^t$ and decreases the likelihood of dispreferred (unsafe) response $y_l^t$ for the target language, which is computed by:
\begin{equation}\small
\begin{split}
    \nabla_\theta \,\text{RG}^{t}(\theta)=&
\frac{1}{|y_w^t|} \,\nabla_\theta \,\log \pi_\theta \bigl(y_w^t\mid x^t\bigr)\\
&-
\frac{1}{|y_l^t|} \,\nabla_\theta \,\log \pi_\theta \bigl(y_l^t\mid x^t\bigr)
\end{split}
\end{equation}
And we have $w_{\theta} = \beta\,\text{RG}^{t}(\theta) - \text{RG}^{d}$, which compares the reward gap between the target language $\beta\,\text{RG}^t$ and the dominant language $\text{RG}^d$. This weight enables the model to adjust both the magnitude and direction of its gradient updates, while the extent of gradient descent is not dictated by the model’s likelihood on the dataset. Thus, $\text{RG}^d$ effectively sets the goal for how strongly the model should discriminate between $y_w^t$ and $y_l^t$ in target languages. Please refer to Appendix \ref{app:grad} for the derivation and detailed discussions.

\textbf{\begin{table*}
\centering
\setlength{\extrarowheight}{0pt}
\resizebox{\linewidth}{!}{
\begin{tabular}{lccccccc|cccccccc|c}
\toprule
\textbf{}  & \multicolumn{7}{c|}{\textbf{MultiJail}} & \multicolumn{8}{c|}{\textbf{AdvBench-X}} & \multicolumn{1}{c}{\textbf{CSRT}} \\
& \textbf{En} & \textbf{Zh} & \textbf{Ko} & \textbf{Ar} & \textbf{Bn} & \textbf{Sw} & \textbf{AVG.} & \textbf{En} & \textbf{Zh} & \textbf{Jp} & \textbf{Ko} & \textbf{Ar} & \textbf{Bn} & \textbf{Sw} & \textbf{AVG.} & \textbf{-} \\
\midrule
\textbf{LLaMA-3.1} & 14.60 & 20.32 & 52.38 & 16.83 & 49.52 & 37.78 & 31.91 & 1.54	&12.5	&17.89	&19.23	&6.15	&40.12	&48.56 &20.86 &18.10\\
\midrule
SFT &12.70	&9.84	&31.43	&8.57	&31.75	&39.37	&22.28 &5.19	&1.73	&2.31	&10.38	&3.08	&18.23	&17.27	&8.31 &13.65 \\
DPO &6.35	&3.17	&15.87	&2.54	&22.86	&37.14	&14.65 &0.77	&1.15	&2.88	&5.58	&\underline{0.38}	&8.83	&18.23	&5.40 &5.71 \\
IPO &7.62	&5.08	&24.44	&2.22	&36.51	&38.73	&19.10 &0.38	&0.77	&3.65	&8.85	&0.96	&10.36	&21.88	&6.69 &\underline{3.49} \\
rDPO &15.24	&14.13	&44.29	&18.73	&50.79	&56.83	&33.34 &6.35	&5.77	&3.85	&11.54	&8.08	&60.65	&56.62	&21.84 &11.43 \\
CPO &22.85	&41.26	&29.21	&38.10	&66.98	&66.98	&44.23 &1.35	&2.69	&3.85	&5.78	&1.35	&20.96	&29.23	&9.32 &19.37 \\
KTO &\underline{4.76}	&6.67	&21.59	&4.76	&30.79	&42.86	&18.57 &0.58	&0.96	&3.27	&8.46	&1.92	&11.35	&22.84	&7.05 &7.31 \\
ORPO &9.52	&\underline{2.86}	&\underline{15.24}	&\textbf{1.27}	&\underline{18.73}	&\underline{21.27}	&\underline{11.48} &\underline{0.19}	&\textbf{0.00}	&\textbf{0.19}	&\textbf{1.35}	&0.58	&11.54	&\underline{10.75}	&\underline{3.51} &3.91 \\
R-DPO &10.16	&14.29	&35.87	&9.84	&42.22	&46.67	&26.51 &3.85	&3.27	&22.31	&3.27	&5.19	&\underline{7.49}	&54.32	&14.24 &11.43 \\
SimPO &9.21	&8.25	&30.48	&7.30	&40.63	&42.22	&23.02 &5.77	&3.46	&11.73	&17.69	&5.19	&28.94	&21.25	&13.43 &7.62 \\
\midrule
\rowcolor{gray!20} MPO (Ours) & \textbf{2.22}	&\textbf{0.95}	&\textbf{4.76}	&\underline{1.90}	&\textbf{12.38}	&\textbf{10.79}	&\textbf{5.98} &\textbf{0.00}	&\underline{0.19}	&\underline{0.38}	&\underline{2.88}	&\textbf{0.00}	&\textbf{7.10}	&\textbf{5.37}	&\textbf{2.27} &\textbf{1.59}\\
\midrule
\midrule
\textbf{Gemma-2} & 2.54	&9.52	&14.61	&4.13	&20.32	&14.60	&10.95 &0.96	&1.15	&3.08	&5.00	&3.85	&6.72	&5.18	&3.71 &4.76\\
\midrule
SFT &2.86	&\underline{4.44}	&13.02	&4.76	&23.17	&12.38	&10.11 &\textbf{0.19}	&\textbf{0.77}	&1.92	&4.42	&\underline{2.50}	&\underline{5.00}	&4.22	&2.72 & 5.74\\
DPO &\underline{2.23}	&7.30	&10.79	&6.35	&23.82	&13.33	&10.64 &0.38	&1.73	&1.54	&3.46	&3.08	&5.03	&\underline{3.84}	&2.72 &5.71 \\
IPO &2.86	&8.89	&16.19	&5.08	&18.41	&14.92	&11.06 &0.77	&1.54	&2.50	&4.42	&3.65	&8.25	&5.18	&3.76 &6.37\\
rDPO &2.54	&8.25	&14.92	&6.35	&20.61	&14.92	&11.27 &0.96	&1.15	&3.27	&4.62	&3.27	&8.45	&5.18	&3.84 &7.62 \\
CPO &3.17	&6.67	&\underline{8.57}	&4.13	&19.68	&13.65	&9.31 &0.38	&1.15	&1.54	&3.85	&4.04	&6.53	&5.57	&3.29 &5.71 \\
KTO &2.23	&6.67	&13.97	&\textbf{3.49}	&20.95	&14.92	&10.37 &0.58	&1.15	&1.92	&4.22	&3.08	&6.14	&4.22	&3.04 &\underline{4.78}\\
ORPO &3.17	&6.03	&10.16	&5.71	&\underline{17.14}	&\underline{10.48}	&\underline{8.78} &0.38	&1.54	&\underline{0.96}	&\underline{2.88}	&\textbf{2.12}	&5.84	&4.26	&\underline{2.57} &6.67 \\
R-DPO &3.81	&7.62	&12.70	&6.35	&28.25	&13.97	&12.12 &0.58	&1.92	&4.42	&4.81	&3.46	&7.68	&4.80	&3.95 &6.03 \\
SimPO &2.54	&8.57	&15.56	&4.44	&20.95	&15.87	&11.32 &0.58	&1.35	&2.69	&4.42	&3.46	&7.10	&4.61	&3.46 &6.67 \\
\midrule
\rowcolor{gray!20} MPO (Ours) & \textbf{0.63}	&\textbf{4.76}	&\textbf{6.98}	&\underline{3.81}	&\textbf{16.51}	&\textbf{7.94}	&\textbf{6.77} &\underline{0.38}	&\underline{0.96}	&\textbf{0.19}	&\textbf{2.50}	&2.69	&\textbf{4.22}	&\textbf{2.88}	&\textbf{1.97} &\textbf{1.90}\\
\bottomrule
\end{tabular}
}
\caption{Detailed results on three multilingual safety benchmarks are presented. The evaluation metric used is the Attack Success Rate (ASR), where lower values indicate better performance. The best results achieved by our method and baselines are highlighted in bold, while the second-best results are underlined.}
\label{tab:main_exp}
\end{table*}}

\section{Experiments}
\subsection{Experimental Setup}
\paragraph{Models} We use the same three backbones as in \S\ref{subsec:reward_gap} to fully validate the efficacy and scalability of our MPO in safety alignment across languages.

\paragraph{Languages to be Safety Aligned} We select six target languages, reflecting diverse linguistic families and resource levels. The high-resource languages are Chinese (Zh) and Japanese (Jp); the medium-resource languages are Korean (Ko) and Arabic (Ar); and the low-resource languages are Bengali (Bn) and Swahili (Sw). For LLaMA-3.1-8B-Instruct and Gemma-2-9B-it, English (En) serves as the dominant language, while that for Qwen2.5-7B-Instruct is En and Zh.

It is crucial to note that these target languages are deemed \emph{out-of-scope} by the official model providers of the three backbones, who stress the importance of additional alignment efforts to guarantee safe and responsible deployment.

\paragraph{Training Data} We sample 100 data points from PKU-SafeRLHF dataset \citep{ji2024pku} and translate them into each target language using the Google Translate API. This leads to that all methods are trained under the same 700 pairs of preference data. Details about the training data can be found in Appendix \ref{app:data}. A comprehensive discussion on the effects of various translation tools and data volumes is provided in \S\ref{subsec:data}.

\paragraph{Benchmarks} To comprehensively measure the efficacy of MPO on various safety scenarios, we employ 3 benchmarks for evaluation, including two multilingual jailbreak datasets: MultiJail \citep{deng2024multilingual} and Advbench-X \citep{yong2023low}, and one code-switch attack dataset: CSRT \citep{yoo2024code}. We use the Attack Success Rate (ASR) as our evaluation metric, calculated according to the evaluation pipeline proposed by \citet{deng2024multilingual} with GPT-4o. Only meaningful refusal responses, excluding unrelated ones, are considered as failed attacks. Please refer to Appendix \ref{app:benchmark} for the detailed description of the evaluation setups.

\paragraph{Baseline Methods}
We compare MPO with supervised finetuning (\textbf{SFT}) \citep{ouyang2022training} and the following preference optimization methods: \textbf{DPO} \citep{rafailov2023direct}, \textbf{IPO} \citep{azar2024general}, \textbf{rDPO} \citep{chowdhury2024provably}, \textbf{CPO} \citep{xu2024cognitive}, \textbf{KTO} \citep{ethayarajh2024kto}, \textbf{ORPO} \citep{hong2024orpo}, \textbf{R-DPO} \citep{park2024disentangling} and \textbf{SimPO} \citep{meng2024simpo}. Please refer to Appendix \ref{app:baseline} for the detailed description of the baseline methods.

\paragraph{Implementation Details}
All the training experiments are conducted on 8 A100 GPUs based on LLaMA-Factory repo \citep{zheng2024llamafactory}. For more details, please refer to the Appendix \ref{app:implement}.

\begin{table}
\centering
\resizebox{\linewidth}{!}{
\begin{tabular}{lcccccc}
\toprule
\textbf{}  & \multicolumn{2}{c}{\textbf{MT-Bench}} & \multicolumn{2}{c}{\textbf{M-MMLU}} & \multicolumn{2}{c}{\textbf{MGSM}} \\
\cmidrule(lr){2-3} \cmidrule(lr){4-5} \cmidrule(lr){6-7}
& \textbf{En} & \textbf{Mul.} & \textbf{En} & \textbf{Mul.} & \textbf{En} & \textbf{Mul.} \\
\midrule
LLaMA-3.1 &7.31	&4.81	&67.70	&45.35	&88.00	&40.13 \\
\rowcolor{gray!20} + MPO & 7.25	&4.92	&67.10	&44.67	&88.00	&44.67 \\
\midrule
Gemma-2 &7.71 &6.60	&73.40 &55.97	&90.00 &72.93 \\
\rowcolor{gray!20} + MPO &7.83 &6.63	&73.40 &55.92	&90.80 &74.80 \\
\bottomrule
\end{tabular}
}
\caption{Results of the multilingual utility evaluation. En denotes the performance of the dominant language, while Mul. represents the average performance across six target languages: Zh, Jp, Ar, Ko, Bn and Sw.}
\label{tab:utility}
\end{table}

\begin{table*}
\centering
\setlength{\extrarowheight}{0pt}
\resizebox{\linewidth}{!}{
\begin{tabular}{lccccccc|cccccccc}
\toprule
\textbf{}  & \multicolumn{7}{c|}{\textbf{MultiJail}} & \multicolumn{8}{c}{\textbf{MT-Bench}} \\
& \textbf{En} & \textbf{Zh} & \textbf{Ko} & \textbf{Ar} & \textbf{Bn} & \textbf{Sw} & \textbf{AVG.} & \textbf{En} & \textbf{Zh} & \textbf{Jp} & \textbf{Ko} & \textbf{Ar} & \textbf{Bn} & \textbf{Sw} & \textbf{AVG.} \\
\midrule
\rowcolor{gray!20} MPO & 2.22	&0.95	&4.76	&1.90	&12.38	&10.79	&5.98 & 7.25	&5.32	&5.26	&5.44	&5.38	&4.11	&4.01	&5.25 \\
\midrule
w/o Retain &2.23	&0.63	&1.90	&0.95	&10.16	&13.33	&\textbf{4.87}	&7.19	&5.09	&4.41	&5.27	&4.79	&3.46	&3.79	&4.86 \\
w/ KL &14.60	&22.54	&58.73	&22.54	&58.10	&75.87	&42.06 &7.41	&5.33	&5.16	&5.58	&5.38	&4.28	&4.11	&5.32 \\
w/o LN &17.78	&26.67	&57.46	&26.67	&65.08	&77.14	&45.13 &7.41	&5.61	&5.29	&5.53	&5.38	&4.31	&4.31	&\textbf{5.41} \\
\bottomrule
\end{tabular}
}
\caption{Ablation results on the key components of MPO. The best results are highlighted in bold.}
\label{tab:other_abl}
\end{table*}

\subsection{Overall Evaluation}

Table \ref{tab:main_exp} demonstrates the performance comparison of MPO and baselines based on LLaMA-3.1-8B-Instruct and Gemma-2-9B-it. Please refer to Appendix \ref{app:main_exp_qwen} for more results on Qwen2.5-7B-Instruct. From the results across all backbones, we have drawn the following key insights:

\paragraph{MPO exhibits robust and consistent performance across various benchmarks and backbone models.} It consistently surpasses all preference learning methods across three backbone LLMs and benchmarks, highlighting its outstanding safety alignment capabilities and scalability.

\paragraph{MPO excels in low-resource languages.} Existing baseline methods often exhibit biased performance, disproportionately benefiting high-resource languages (e.g., Zh and Jp) and those where the model already demonstrates strong safety alignment (e.g., Ar). In contrast, MPO achieves comprehensive and significant improvements, particularly in low-resource languages (e.g., Bn and Sw). This highlights the effectiveness of leveraging high-quality internal safety alignment signals instead of relying exclusively on uneven preference data.

\paragraph{MPO maintains multilingual utility.}
Multilingual safety alignment should not compromise the model’s multilingual general utility. Thus, we evaluate the resulting model across three key dimensions: (1) World Knowledge: M-MMLU \citep{hendrycks2021measuring}, (2) Reasoning: MGSM \citep{shi2023language}, and (3) Multi-turn Instruction-Following: MT-Bench \citep{zheng2023judging}. The results in Table \ref{tab:utility} show that MPO consistently maintains the general utility of both the dominant and target languages. For detailed results, evaluation settings and the comparison with baseline methods, please refer to Appendix \ref{app:utility}.

\begin{figure}
\centering
\includegraphics[width=1.0\columnwidth]{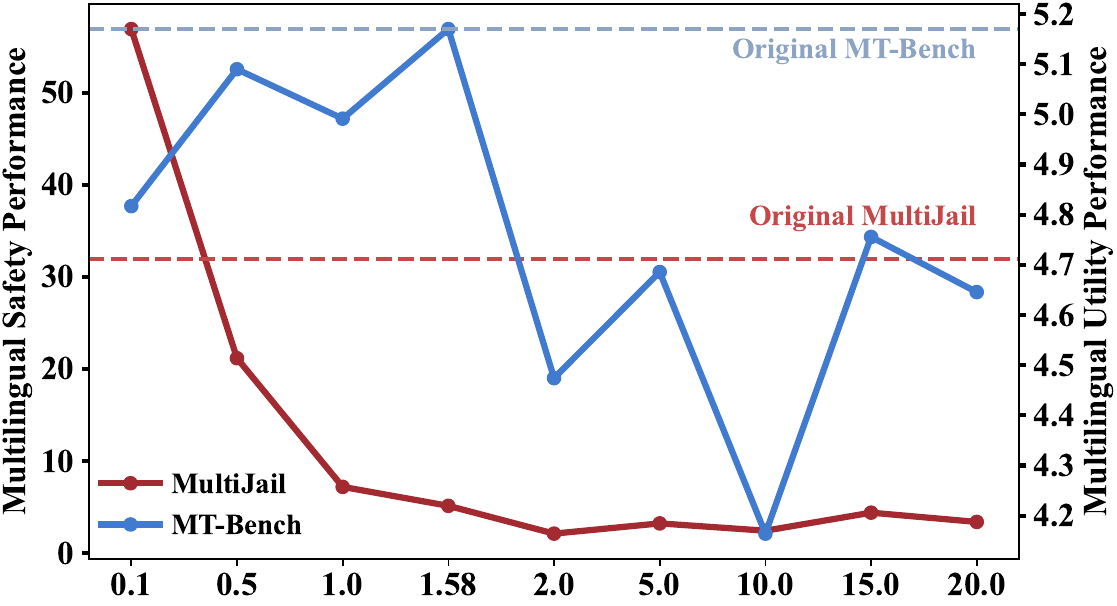}
\caption{The results of replacing the dominant language reward gap with a fixed value on multilingual safety and general utility performance.}
\label{fig:fix_value}
\end{figure}

\section{Analysis and Discussions}
\label{sec:analysis}
In this section, we offer a comprehensive analysis of MPO from: 
(1) ablation studies (\S\ref{subsec:ablation}), (2) the influence of preference data quality and quantity (\S\ref{subsec:data}), and (3) the rewards, representations, and case visualizations of the resulting model (\S\ref{subsec:case}). Unless stated otherwise, all analysis are conducted using the LLaMA-3.1 backbone.

\subsection{Ablation Study}
\label{subsec:ablation}

\begin{figure*}
\centering
\subfigure[Impact of Data Quality.]{
\includegraphics[width=1.0\columnwidth]{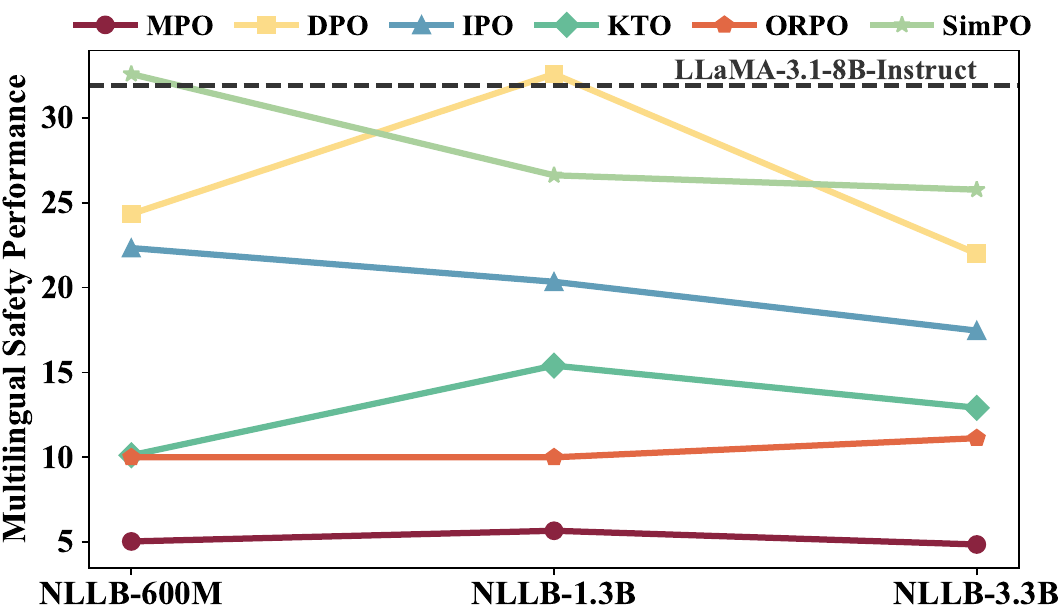}
\label{subfig:data_quality}
}
\subfigure[Impact of Data Quantity.]{
\includegraphics[width=1.0\columnwidth]{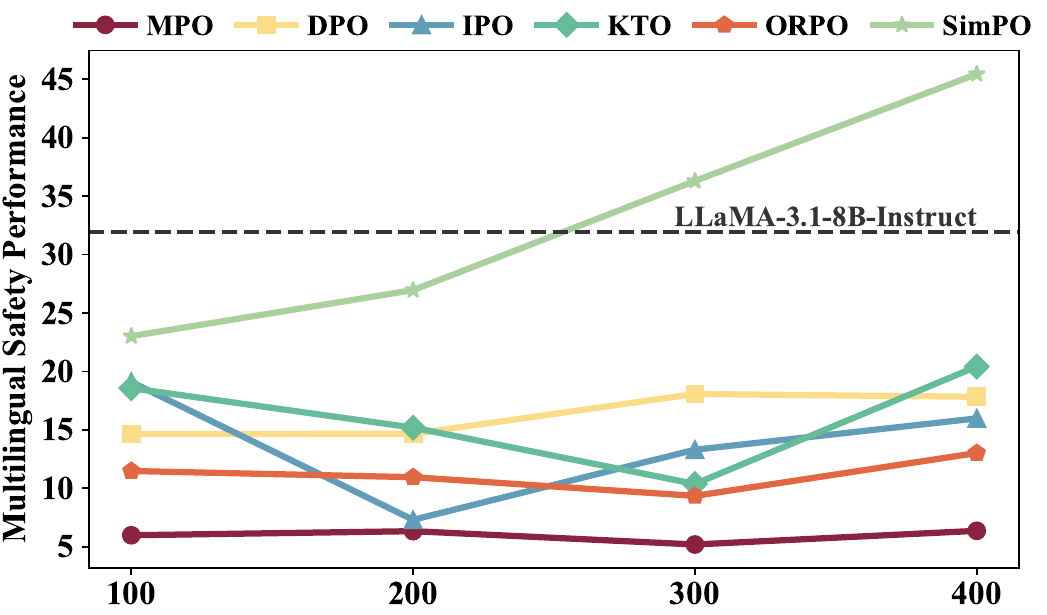}
\label{subfig:data_quantity}
}
\caption{Impact of the preference data. (a) Multilingual safety performance on MultiJail with varied data quality. (b) Multilingual safety performance on MultiJail with varied data size.}
\label{t5_scale}
\end{figure*}

\paragraph{Effect of Reward Gap from the Dominant Language as the Supervision Signal} To assess the effectiveness of using the reward gap from the dominant language as an alignment objective, we conduct ablation experiments where we replace it with either a \emph{fixed constant} or the \emph{reward gap of other languages}. We also compare MPO against \emph{recent cross-lingual transfer methods}.

Figure \ref{fig:fix_value} shows the impact of replacing the dominant language reward gap with a fixed value (0.1–20) on multilingual safety and general utility. While increasing the constant enhances safety performance, it significantly degrades general utility due to excessive parameter shifts, leading to model collapse despite retention constraints. Notably, setting the constant to 1.58 (the training set’s average reward gap of the dominant language) yields limited gains, highlighting the superiority of the fine-grained instance-level supervision in our MPO over coarse-grained dataset-level alignment. Please see Appendix \ref{app:ablation} for more details.

\begin{table}
\centering
\resizebox{\linewidth}{!}{
\begin{tabular}{lccccccc}
\toprule
\textbf{}  & \multicolumn{7}{c}{\textbf{MultiJail}} \\
\cmidrule(lr){2-8}
& \textbf{En} & \textbf{Zh} & \textbf{Ko} & \textbf{Ar} & \textbf{Bn} & \textbf{Sw} & \textbf{AVG.}\\
\midrule
LLaMA-3.1 & 14.60 & 20.32 & 52.38 & 16.83 & 49.52 & 37.78 & 31.91 \\
\midrule
Align with Ar &6.98	&6.67	&20.00	&4.13	&17.78	&46.35	&16.99 \\
Align with Bn &35.56	&41.91	&72.07	&51.75	&63.49	&84.44	&58.20 \\
Align with Sw &20.63	&30.16	&53.97	&26.35	&53.97	&81.90	&44.50 \\
\midrule
\rowcolor{gray!20} MPO &\textbf{2.22}	&\textbf{0.95}	&\textbf{4.76}	&\textbf{1.90}	&\textbf{12.38}	&\textbf{10.79}	&\textbf{5.98} \\
\bottomrule
\end{tabular}
}
\caption{Multilingual safety performance when replacing reward gap with that from Ar, Bn and Sw as the supervision signal. The evaluation metric is the Attack Success Rate (ASR), where lower values indicate better performance. The best results are highlighted in bold.}
\label{tab:other_lang}
\end{table}

Table \ref{tab:other_lang} further shows that using the reward gap of a target language as the alignment objective fails to yield meaningful safety improvements. Even when selecting the second-best safety-performing language (Ar) or low-resource languages (Sw, Bn), no effective multilingual safety enhancement is observed. This reinforces that the dominant language’s reward gap provides a more reliable and high-quality supervision signal. For Qwen2.5, although it is a bilingual LLM with both Chinese and English as dominant languages, we find that using Chinese as the alignment target leads to better safety alignment performance compared to using English. Detailed results supporting this observation are provided in Appendix \ref{app:main_exp_qwen}, Table \ref{tab:main_exp_qwen}.

Table \ref{tab:cross_lingual} in Appendix \ref{app:ablation} compares MPO with state-of-the-art cross-lingual transfer methods, which align multilingual safety by either aligning multilingual representations: CLA \citep{li2024improving} and \textsc{Lens} \citep{zhao2024lens}, or distilling knowledge from the dominant language: SDRRL \citep{zhang2024enhancing}. MPO consistently outperforms these methods, maintaining strong multilingual safety alignment. This further highlights the advantage of leveraging the dominant language’s reward gap as a fine-grained supervision signal.

\paragraph{Effect of Other Components in MPO} We further analyze the effect of other key components in Table \ref{tab:other_abl}. Removing the Retain component in Eq. (\ref{eq:rep_cons}) leads to a significant drop in multilingual utility, demonstrating its efficacy in preserving cross-lingual robustness. Introducing a KL-divergence-based constraint imposes a strong regularization that restricts the alignment of reward gap distributions across different languages, limiting the flexibility of MPO in adapting to multilingual safety preferences. Finally, removing length normalization (LN) in reward gap computation results in biased reward gap values, particularly in safety scenario that unsafe responses are often longer than safe ones, highlighting that LN effectively mitigates length-induced bias and facilitates more stable multilingual safety alignment.

\subsection{The Impact of Preference Data}
\label{subsec:data}

\paragraph{Impact of Data Quality} To evaluate the robustness of MPO across different levels of multilingual preference data quality, we employ three versions of the dataset obtained using three NLLB \citep{costa2022no} translation models of varying sizes: NLLB-600M, NLLB-1.3B, and NLLB-3.3B. These models represent a progressive improvement in translation quality, with the largest model generally producing more accurate translations. Results are shown in Figure \ref{subfig:data_quality}.

Baselines show considerable performance variations across different data quality levels, struggling to maintain stable safety alignment—even when trained on the highest-quality preference data (NLLB-3.3B). This underscores the challenges that noisy multilingual data pose for existing alignment methods. In contrast, MPO consistently delivers the best results across all data quality levels, demonstrating its stability and resilience to data noise. This validates the effectiveness of leveraging the reward gap in the dominant language as a source of high-quality supervision.

Further, recent studies explore LLMs themselves to generate multilingual preference data, rather than relying on external translation tools \citep{she2024mapo,yang2024language}. MPO consistently achieves the best multilingual safety alignment results data sources, demonstrating its robustness to variations in preference data. Please refer to Appendix \ref{app:data_source} for detailed results and analysis.

\paragraph{Impact of Data Quantity} Figure \ref{subfig:data_quantity} compares MPO with baseline methods across varying dataset sizes, with the x-axis representing the number of preference samples per language. MPO maintains stable performance across different data volumes, consistently outperforming baselines. However, all methods, including MPO, exhibit diminishing returns as data increases, with baseline performance even degrading with excessive data. This highlights that enhancing supervision signal quality is far more effective than simply increasing data volume, aligning with broader LLM post-training trends \citep{zhou2023lima,cao2024towards,guo2025deepseek,ye2025limo}.

\begin{figure}
\centering
\includegraphics[width=1.0\columnwidth]{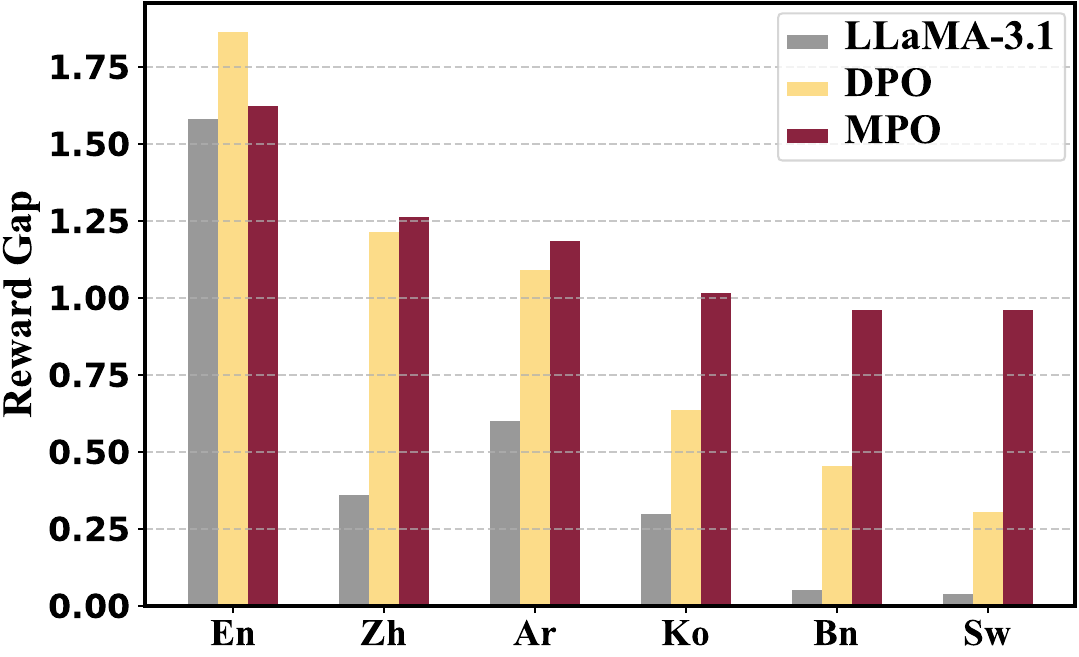}
\caption{Reward gap across languages for the original backbone and those safety aligned by MPO and DPO.}
\label{fig:reward_gap}
\end{figure}

\subsection{Visualization Analysis}
\label{subsec:case}
\begin{figure}
\centering
\includegraphics[width=1.0\columnwidth]{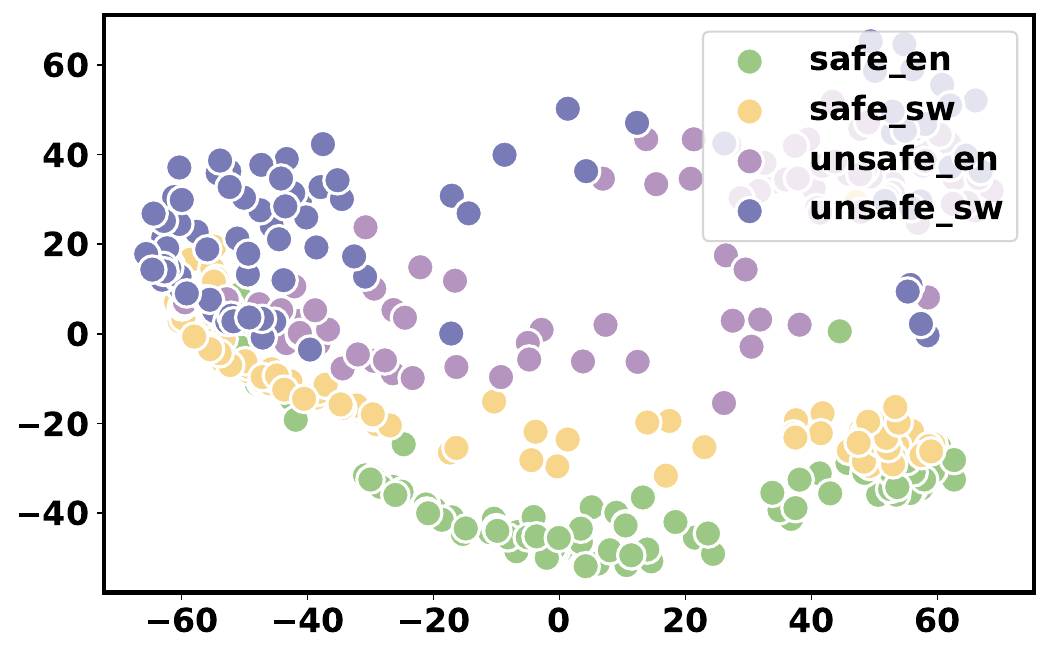}
\caption{The visualization of multilingual representations for English and Swahili.}
\label{fig:visual_rep}
\end{figure}

To better illustrate the impact of MPO on multilingual safety alignment, we visualize changes in the reward gap and the model’s internal representation space. In Figure \ref{fig:reward_gap}, MPO consistently achieves a higher reward gap than DPO across all languages. Notably, it significantly improves low-resource languages such as Swahili and Bengali, reducing the performance gap with English. Further, the visualization of the model’s representation space in Figure \ref{fig:visual_rep}, shows that MPO enables a clearer distinction between safe and unsafe responses in the target language Sw. This suggests that MPO enhances the model’s ability to differentiate safety-critical responses, reinforcing its effectiveness in multilingual safety alignment. Please refer to Appendix \ref{app:visual} for more visualization results.

\section{Related Works}

\paragraph{Multilingual Safety Vulnerability}
Recent studies have exposed risks in the multilingual safety of LLMs, underscoring the need for multilingual safety alignment \citep{qin2024multilingual,li2024xtrust,gupta2024walledeval,kanepajs2024towards,verma2025hidden}. One line of approaches translate harmful prompts from high-resource to low-resource languages to assess safety \citep{yong2023low,deng2024multilingual,xu2024cognitive,shen2024language,li2024cross,wang-etal-2024-languages,poppi2024towards}, as seen in \citet{deng2024multilingual}, which manually translated 315 English safety prompts \citep{ganguli2022red} into nine languages. Others evaluate multilingual safety using code-switching, embedding multiple languages within the same harmful input \citep{gutierrez1999language,yoo2024code,song2024multilingual,upadhayay2024sandwich}.

While these works have established a solid testbed for multilingual safety in LLMs, they have yet to introduce effective solutions to the existing challenges in this domain.

\paragraph{Safety Alignment Technique} DPO \citep{rafailov2023direct} has emerged as a widely adopted offline preference learning method for aligning LLMs with human safety principles and values. In addition to DPO, various preference optimization objectives have been introduced. Ranking-based objectives enable comparisons among more than two instances \citep{dong2023raft,yuan2023rrhf,liu2024lipo,song2024preference}. IPO \citep{azar2024general} mitigates the overfitting issues inherent in DPO, while KTO \citep{ethayarajh2024kto} addresses preference optimization in non-pairwise data settings. Meanwhile, ORPO \citep{hong2024orpo} and SimPO \citep{meng2024simpo} seek to remove reliance on a reference model.

Our proposed MPO stands out from existing methods in that we seek multilingual supervision signals from the \emph{internal} reward gap of the LLMs, which specifically addresses the challenge of uneven data quality in multilingual safety alignment and offers fresh insights and new opportunities for achieving effective multilingual safety alignment.

\section{Conclusion}
In this paper, we introduce MPO, a novel approach to multilingual safety alignment that leverages the reward gap of the dominant language as a high-quality supervision signal. MPO directly minimizes the discrepancy of reward gap across different languages to transfer safety alignment effectively.
Experiments on LLaMA-3.1, Gemma-2, and Qwen2.5 confirm that MPO outperforms existing methods in multilingual safety alignment without compromising general multilingual utility. Further analysis shows that MPO remains robust across varying data qualities and sources, reinforcing the superiority of the dominant language’s reward gap as a scalable alignment signal. These results establish MPO as a practical and effective solution for deploying multilingual-safe LLMs.

\section*{Limitations}
This work has several limitations that provide directions for future research. Due to computational constraints, we conduct experiments on mid-scale models and did not extend our evaluation to larger-scale ones such as 32B even 72B LLMs. Future work should explore whether MPO scales effectively with larger models and whether its advantages persist at greater parameter sizes.

Additionally, we have focused exclusively on the application of MPO to multilingual safety alignment. However, there are more challenging and diverse alignment tasks that could be explored in the future, particularly those involving multicultural value alignment \citep{sorensen2024roadmap,yao2024value,cahyawijaya2024high}. As multilingual safety alignment is only one aspect of broader ethical considerations, future work could extend the current methodology to tackle these value alignment challenges, ensuring models respect different cultural norms and ethical standards across regions.

Furthermore, given that safety guidelines are universal principles that users across various linguistic and cultural regions must adhere to, as emphasized in OpenAI \citep{openai2024use} and Meta’s user guidelines \citep{meta2024safety}, it is reasonable to transfer the safety alignment of the dominant language to other languages. This idea has proven effective in our experiments, and we believe it could be validated in broader multilingual tasks in the future, particularly those that are language-agnostic, such as general problem-solving skills \citep{hu2024large,zhang2024getting,huang20241+,wang2024sharing}. Future work could explore these areas and broaden the scope of multilingual model evaluation, to ensure that advanced AI technologies are universally applicable and can promote responsible and ethical AI development on a global scale. We hope the research community continues to push forward in advancing these technologies and facilitating their global adoption.

\section*{Ethical Considerations}
This work is conducted solely for academic research purposes and aims to address multilingual safety risks in large language models (LLMs). The primary goal of our study is to improve the robustness and consistency of LLMs across different languages, ensuring that they adhere to established safety principles regardless of linguistic variations. We acknowledge that multilingual safety alignment is a complex challenge, and our research does not aim to impose any specific cultural or ethical standards on diverse linguistic communities. Instead, our approach focuses on enhancing model consistency in following universally recognized safety guidelines, as outlined in user policies of major AI developers such as OpenAI and Meta. By ensuring equitable safety alignment across languages, we seek to mitigate risks associated with uneven safety performance in LLMs and reduce potential harm in lower-resource languages.

In conclusion, we aim to contribute to the development of fair, transparent, and globally applicable AI systems that align with responsible AI deployment principles. We encourage further community-driven research to refine multilingual safety alignment and promote the ethical and safe application of AI technologies worldwide.

\section*{Acknowledgments}
We thank the anonymous reviewers for their comments and suggestions. This work was supported by the National Key RD Program of China via grant 2021YFF0901602, the National Natural Science Foundation of China (NSFC) via grant 62176078, the Fundamental Research Funds for the Central Universities and the Singapore Ministry of Education (MOE) Academic Research Fund (AcRF) Tier 1 grant (No. MSS24C012).
% Bibliography entries for the entire Anthology, followed by custom entries
%\bibliography{anthology,custom}
% Custom bibliography entries only
\bibliography{custom}

\newpage

\appendix

\begin{table}
\centering
\resizebox{\linewidth}{!}{
\begin{tabular}{cccccccc}
\toprule 
& & \textbf{En} & \textbf{Zh} & \textbf{Ko} & \textbf{Ar} & \textbf{Bn} & \textbf{Sw} \\
\midrule
\multirow{2}{*}{\textbf{LLaMA-3.1}} &\textbf{RG}$\uparrow$ & \textbf{27.02} &25.09 &22.82 &27.85 &32.21 &29.77 \\
&\textbf{ASR}$\downarrow$ &\textbf{9.00}	&22.00	&50.00	&15.00	&55.00	&57.00 \\
\midrule
\multirow{2}{*}{\textbf{Gemma-2}} &\textbf{RG}$\uparrow$ &\textbf{214.57} &63.92 &45.54 &88.46 &80.62 &96.60 \\
&\textbf{ASR}$\downarrow$ &\textbf{0.00}	&9.00	&14.00	&4.00	&24.00	&26.00 \\
\midrule
\multirow{2}{*}{\textbf{Qwen-2.5}} &\textbf{RG}$\uparrow$ &\textbf{13.84} &\textbf{12.31} &9.77 &19.76 &5.04 &23.77 \\
&\textbf{ASR}$\downarrow$ &\textbf{13.00} &\textbf{9.00}	&21.00	&20.00	&69.00	&98.00 \\
\bottomrule 
\end{tabular}
}
\caption{Results of reward gap (calculated by Eq. \ref{eq:dpo_rg}) and safety performance across six languages. The evaluation metric used for safety is the Attack Success Rate (ASR), where lower values indicate better performance. Results of the dominant languages are highlighted in bold.}
\label{tab:reward_gap_dpo}
\end{table}

\section{Further Discussion on Reward Gap}
\label{app:rg_discuss}
We have already discussed in \S\ref{subsec:simpo} that using Eq. (\ref{eq:simpo_rg}) within SimPO \citep{meng2024simpo} to compute the reward gap is more reasonable compared to Eq. (\ref{eq:dpo_rg}) in DPO \citep{rafailov2023direct}. Additionally, in \S\ref{subsec:reward_gap}, we provide an intuitive demonstration of the advantages of using Eq. (\ref{eq:simpo_rg}) for reward gap calculation. Here, we conduct a more in-depth analysis to illustrate the limitations of Eq. (\ref{eq:dpo_rg}).

Table \ref{tab:reward_gap_dpo} presents the results of computing the reward gap using Eq. (\ref{eq:dpo_rg}) across three different backbone models. $\beta$ is set to 1.0 and the base version of these backbones are adopted as the reference models. The dataset used for this evaluation remains consistent with that in \S\ref{subsec:reward_gap}. However, the results indicate that the computed reward gap fails to accurately reflect the model’s safety performance across different languages. We attribute this discrepancy to the following three key reasons:

(1) \textbf{Inference-Training Objective Mismatch}: The reward formulation in Eq. (\ref{eq:dpo_rg}) is derived from the implicit reward used during the training phase, but it does not directly align with the log-likelihood objective that governs inference \citep{meng2024simpo}. As a result, the reward gap computed with Eq. (\ref{eq:dpo_rg}) may not faithfully capture the model’s actual generation behavior, leading to misleading safety performance evaluations.

(2) \textbf{Bias in the Reference Model}: Ideally, the reference model used for computing the reward gap in a preference-optimized model should be the supervised fine-tuned (SFT) model from the previous training stage, rather than the base model \citep{ouyang2022training,rafailov2023direct}. However, model providers do not publicly release this intermediate SFT model, making it difficult to obtain an accurate reference. As a result, using the base model as the reference introduces bias \citep{hong2024orpo,wu2024alpha}, further compromising the reliability of Eq. (\ref{eq:dpo_rg}) in assessing safety performance.

(3) \textbf{Length Bias Effects}: Eq. (\ref{eq:dpo_rg}) does not incorporate length normalization, making it susceptible to biases introduced by response length disparities \citep{meng2024simpo,kim2024rethinking}. Empirically, unsafe responses tend to be longer due to the presence of explicit harmful content, while safe responses are often concise refusals. This discrepancy skews the reward gap calculations, causing inconsistencies in cross-linguistic safety evaluations.

These limitations collectively suggest that Eq. (\ref{eq:dpo_rg}) from DPO \citep{rafailov2023direct} is not a reliable metric for evaluating safety differences across languages. In contrast, Eq. (\ref{eq:simpo_rg}) from SimPO \citep{meng2024simpo} mitigates these issues by normalizing the log-likelihood with sequence length, ensuring a more accurate measure of safety performance.

\section{Further Discussion on MPO}
\label{app:MPO_discuss}

Here we conduct further explanation of how reward gap of the dominant language $\text{RG}^d$ influences the learning of $y_w^t$ and $y_l^t$ for different languages.

Recall that:
{\small\begin{equation}
\text{RG}^d =
\frac{1}{\lvert y_w^d\rvert}\,\log \pi_{\text{ref}}\bigl(y_w^d\!\mid x^d\bigr)
-
\frac{1}{\lvert y_l^d\rvert}\,\log \pi_{\text{ref}}\bigl(y_l^d\!\mid x^d\bigr), \notag
\end{equation}}
where $\pi_{\text{ref}}$ is a reference policy (or model). This quantity, $\text{RG}^d$, is constant with respect to the trainable parameters $\theta$ (because it depends on the reference model). However, it plays an important role in shaping how $\theta$ is learned for $y_w^t$ and $y_l^t$.

\paragraph{Target Gap} The difference $\beta \,\text{RG}^t - \text{RG}^d$ appears inside the loss function. Because $\text{RG}^d$ is subtracted from $\beta\,\text{RG}^t$, it effectively sets a target ``goal'' in log probabilities that the model $\pi_\theta$ should achieve between the winning (safe) candidate $y_w^t$ and the losing (unsafe) candidate $y_l^t$.

\paragraph{Penalty for Reward Signal} The term $\beta\,\text{RG}^t - \text{RG}^d$ penalizes deviations of $\beta\,\text{RG}^t$ from $\text{RG}^d$. Intuitively, if $\beta\,\text{RG}^t$ is not aligned with $\text{RG}^d$, the loss increases, thus signaling the training process that $\pi_\theta$ is not matching the reference gap.

\paragraph{Alignment with Reference Behavior} Because $\text{RG}^d$ comes from $\pi_{\text{ref}}$, one can interpret it as how strongly the reference policy prefers its ``winning'' candidate $y_w^d$ over its ``losing'' candidate $y_l^d$. By forcing $\text{RG}^t$ from $\pi_\theta$ to approximate $\text{RG}^d$, the training encourages $\pi_\theta$ to mimic or at least stay consistent with that preference structure, though for potentially different $x^t$ and $y_w^t$, $y_l^t$.

\paragraph{Effect on $y_w^t$ and $y_l^t$ Learning}

For $y_w^t$: If the reference gap indicates a high preference for a corresponding ``winning'' candidate $y_w^d$, then during training, the model sees a stronger incentive to increase $\log \pi_\theta(y_w^t\mid x^t)$ (since that helps match the overall gap).

For $y_l^t$: The model similarly sees a signal to decrease $\log \pi_\theta(y_l^t\mid x^t)$ (or at least not let it grow too large), in order to keep the difference consistent with $\text{RG}^d$.

In essence, $\text{RG}^d$ provides a reference or target difference in log probabilities that the model $\pi_\theta$ tries to match between $y_w^t$ and $y_l^t$. Although it does not directly update $\theta$ (because it is constant with respect to $\theta$), it influences the loss landscape and hence indirectly guides how $\log \pi_\theta(y_w^t\mid x^t)$ and $\log \pi_\theta(y_l^t\mid x^t)$ are learned.

\section{Gradient Analysis of MPO}
\label{app:grad}

\subsection{Deriving the Gradient of MPO}

Below is a step-by-step derivation of the gradient of the loss function.

{\small\begin{equation}
\label{eq:loss}
\mathcal{L}(\pi_\theta) \;=\; 
\mathbb{E}_{(x, y_w, y_l) \,\sim\, \mathcal{D}} 
\Bigl[ 
   \bigl\|\beta\,\text{RG}^{t} \;-\; \text{RG}^{d}\bigr\|^{2}
\Bigr],
\end{equation}}
where
{\small\begin{equation}
\label{eq:RGt}
\text{RG}^{t} = 
\frac{1}{\lvert y_w^t\rvert}\,\log \pi_\theta \bigl(y_w^t \mid x^t\bigr)
-
\frac{1}{\lvert y_l^t\rvert}\,\log \pi_\theta \bigl(y_l^t \mid x^t\bigr),
\end{equation}}
and
{\small\begin{equation}
\label{eq:RGd}
\text{RG}^{d} 
\;=\; 
\frac{1}{\lvert y_w^d\rvert}\,\log \pi_{\text{ref}}\bigl(y_w^d \mid x^d\bigr)
\;-\;
\frac{1}{\lvert y_l^d\rvert}\,\log \pi_{\text{ref}}\bigl(y_l^d \mid x^d\bigr).
\end{equation}}
Note that \(\text{RG}^{d}\) does not depend on \(\theta\), whereas \(\text{RG}^{t}\) depends on \(\theta\) through \(\log \pi_\theta(\cdot)\).

\paragraph{Rewrite the Loss Function} Define the per-sample loss (ignoring the expectation for a moment) as
\begin{equation}\small
\ell(\theta)
\;=\;
\bigl\|\beta \,\text{RG}^{t}(\theta) \;-\; \text{RG}^{d}\bigr\|^{2}.
\end{equation}
For the gradient derivation, we focus on \(\ell(\theta)\). The overall gradient will then be its expectation w.r.t.\ the data distribution \(\mathcal{D}\).

\paragraph{Introduce an Intermediate Variable}

Let
\begin{equation}\small
z(\theta) 
\;=\; 
\beta \,\text{RG}^{t}(\theta) \;-\; \text{RG}^{d}.
\end{equation}
Hence
\begin{equation}\small
\ell(\theta)
\;=\;
\bigl\|z(\theta)\bigr\|^2.
\end{equation}
If \(\text{RG}^{t}\) is a scalar, \(\|z(\theta)\|^2 = z(\theta)^2\). (For the vector case, one may treat each component in the same way.)

\paragraph{Apply the Chain Rule to \(\ell(\theta)\)}

We have
\begin{equation}\small
\ell(\theta)
\;=\;
\bigl\|z(\theta)\bigr\|^2.
\end{equation}
Taking the gradient w.r.t.\ \(\theta\),
{\small{\begin{align}
\nabla_\theta \,\ell(\theta)
\;&=\;
\nabla_\theta \,\bigl\|z(\theta)\bigr\|^2 \\
&=\;
2\;z(\theta)\;\nabla_\theta \,z(\theta).
\end{align}}}
Recalling 
\begin{equation}\small
z(\theta) 
\;=\; 
\beta \,\text{RG}^{t}(\theta) \;-\; \text{RG}^{d},
\end{equation}
and that \(\text{RG}^{d}\) is a constant w.r.t.\ \(\theta\), we get:
\begin{equation}\small
\nabla_\theta \,z(\theta)
\;=\;
\beta \,\nabla_\theta \,\text{RG}^{t}(\theta).
\end{equation}
Hence,
\begin{equation}\small
\nabla_\theta \,\ell(\theta)
\;=\;
2\,
\Bigl[\beta \,\text{RG}^{t}(\theta) - \text{RG}^{d}\Bigr]
\;\beta\,
\nabla_\theta \,\text{RG}^{t}(\theta).
\end{equation}

\paragraph{Compute \(\nabla_\theta \,\text{RG}^{t}(\theta)\)}

By definition,
\begin{equation}\small
\text{RG}^{t}(\theta)
\;=\;
\frac{1}{|y_w^t|}\,\log \pi_\theta \bigl(y_w^t \mid x^t\bigr)
\;-\;
\frac{1}{|y_l^t|}\,\log \pi_\theta \bigl(y_l^t \mid x^t\bigr).
\end{equation}
Hence,
\begin{equation}\small
\begin{split}
\nabla_\theta \,\text{RG}^{t}(\theta)=
\frac{1}{|y_w^t|} \,\nabla_\theta \,\log \pi_\theta \bigl(y_w^t\mid x^t\bigr)
\\-
\frac{1}{|y_l^t|} \,\nabla_\theta \,\log \pi_\theta \bigl(y_l^t\mid x^t\bigr).
\end{split}
\end{equation}

\paragraph{Combine the Results}

Putting it all together,
\begin{equation}\small
\begin{split}
\nabla_\theta \,\ell(\theta)
\;=\;
2\,
\Bigl[\beta \,\text{RG}^{t}(\theta) - \text{RG}^{d}\Bigr]
\;\\ \beta\,
\Bigl(
  \frac{1}{|y_w^t|}\,\nabla_\theta \,\log \pi_\theta(y_w^t\mid x^t)
  \;-\;
  \frac{1}{|y_l^t|}\,\nabla_\theta \,\log \pi_\theta(y_l^t\mid x^t)
\Bigr).
\end{split}
\end{equation}

\paragraph{The Full Gradient of \(\mathcal{L}(\theta)\)}

Recall the original loss is the \emph{expectation} of \(\ell(\theta)\) over samples \((x,y_w,y_l)\sim\mathcal{D}\). Therefore,
\begin{equation}\small
\label{eq:grad_1}
\begin{split}
\nabla_\theta \,\mathcal{L}(\theta)
\;=\;
2\beta\,\mathbb{E}_{(x, y_w, y_l)\,\sim\,\mathcal{D}}
\Biggl[
   \bigl(\beta\,\text{RG}^{t}(\theta) - \text{RG}^{d}\bigr)
   \\\biggl(
     \frac{1}{|y_w^t|}\,\nabla_\theta \,\log \pi_\theta(y_w^t \mid x^t)
     \;-\;
     \frac{1}{|y_l^t|}\,\nabla_\theta \,\log \pi_\theta(y_l^t \mid x^t)
   \biggr)
\Biggr].
\end{split}
\end{equation}

This completes the derivation of the gradient w.r.t.\ the parameters \(\theta\).

\subsection{Analysis}

Here we explain how $\text{RG}^d$ influences the model’s updates for $y_w^t$ and $y_l^t$ from the gradient view:

\paragraph{Shifts the Gradient Magnitude and Direction} The difference $(\beta\,\text{RG}^t(\theta) - \text{RG}^d)$ multiplies the gradient terms that involve $\log \pi_\theta(y_w^t \mid x^t)$ and $\log \pi_\theta(y_l^t \mid x^t)$. If $\beta\,\text{RG}^t$ is smaller than $\text{RG}^d$, then the difference is negative, which encourages the model to increase $\text{RG}^t$ (e.g., by increasing the probability of $y_w^t$ or decreasing the probability of $y_l^t$) so that it moves toward or surpasses $\text{RG}^d$. If $\beta\,\text{RG}^t$ is larger than $\text{RG}^d$, the difference is positive, so the model is nudged to preserve or even enlarge its current gap, reinforcing the discrimination it has already learned between $y_w^t$ and $y_l^t$.

\paragraph{Controls the Drive to Differentiate $y_w^t$ and $y_l^t$} Because $\text{RG}^t$ involves $\log \pi_\theta(y_w^t)$ and $\log \pi_\theta(y_l^t)$, the difference term $(\beta\,\text{RG}^t(\theta) - \text{RG}^d)$ directly scales how strongly the model updates its parameters to favor $y_w^t$ over $y_l^t$. A larger $\text{RG}^d$ essentially raises the ``bar'' the model is trying to clear; a smaller $\text{RG}^d$ lowers it.

\section{Training Data}
\label{app:data}
To ensure that our multilingual preference data generation process remains on-policy, we adopt a structured approach based on well-established principles in LLM post-training \citep{dubey2024llama}. Specifically, for each English harmful prompt in the PKU-SafeRLHF dataset \citep{ji2024pku}, we first feed it to the model to generate a refusal response, which serves as the preferred response. We then pair this generated refusal with the original dispreferred response from the dataset, forming a preference pair. This ensures that the optimization process remains aligned with the model’s actual behavior, avoiding potential inconsistencies that arise from using static preference data \citep{yuan2024self,chen2024self,rosset2024direct,wu2024self,guo2024direct,zhou2025dreamdpo}.

To extend this preference data to multiple languages, we then translate both the harmful prompts and their paired responses using the Google Translate API. This approach allows us to create multilingual preference data while preserving the preference structure of the original dataset.

\section{Benchmark}
\label{app:benchmark}
We comprehensively measure the efficacy of our MPO on various multilingual safety benchmarks.
\begin{itemize}
    \item \textbf{MultiJail} \citep{deng2024multilingual}: It carefully gather 315 English harmful queries and manually translate them by native speakers into 9 non-English languages, ranging from high-resource to low-resource. 
    \item \textbf{AdvBench-X} \citep{yong2023low}: AdvBench is a set of 500 harmful behaviors formulated as instructions. These behaviors range over the same themes as the harmful strings setting, but the adversary’s goal is to find a single attack string that will cause the model to generate any response that attempts to comply with the instruction, and to do so over as many harmful behaviors as possible. The original English version is also translated manually into target languages of different resource levels.
    \item \textbf{CSRT} \citep{yoo2024code}: It synthesizes code-switching red-teaming queries, combining up to 10 languages, and investigate the safety and multilingual understanding of LLMs.
\end{itemize}
We evaluate multilingual safety alignment using the Attack Success Rate (ASR), following the evaluation pipeline proposed by \citet{deng2024multilingual}, with GPT-4o as the judgment model. The evaluation process consists of the following steps: (1) Translation to English: Since safety alignment performance needs to be assessed across multiple languages, we first translate the model-generated responses from the target language into English using GPT-4o to ensure consistent evaluation. (2) Three-Class Classification: GPT-4o then classifies each response into one of the following categories: Safe (meaningful refusal), Unsafe or Irrelevant. (3) Attack Success Calculation: Responses classified as unsafe or irrelevant are both considered unsuccessful refusals and thus counted as successful attacks when calculating ASR. Only safe refusals are considered failed attacks, contributing to a lower ASR (better safety performance).

It is essential to highlight that the languages targeted for enhancement, as mentioned above, are all within the capability range of GPT-4o, especially given that its official model card \citep{openai2024gpt4o} emphasizes support for low-resource languages such as Swahili (Sw) and Bengali (Bn). This underscores the validity and reliability of the evaluation approach.

\begin{table*}
\centering
\resizebox{\linewidth}{!}{
\begin{tabular}{ccccccccc}
\toprule
\textbf{Method} & \textbf{Objective} & \textbf{Hyperparameter} \\
\midrule
DPO \citep{rafailov2023direct} & $-\log \sigma \left( \beta \log \frac{\pi_\theta(y_w|x)}{\pi_{\text{ref}}(y_w|x)} - \beta \log \frac{\pi_\theta(y_l|x)}{\pi_{\text{ref}}(y_l|x)}\right)$ & $\beta \in [0.01, 0.05, 0.1]$ \\ \midrule
IPO \citep{azar2024general} & $ \left( \log \frac{\pi_\theta(y_w|x)}{\pi_{\text{ref}}(y_w|x)} - \log \frac{\pi_\theta(y_l|x)}{\pi_{\text{ref}}(y_l|x)} - \frac{1}{2\tau} \right)^2$ & $\tau \in [0.01, 0.1, 0.5, 1.0]$ \\  \midrule 
\multirow{2}{*}{rDPO \citep{chowdhury2024provably}} & $\frac{(1-\epsilon)\mathcal{L}(\theta,x,y_w,y_l)-\epsilon \mathcal{L}(\theta,x,y_l,y_w)}{1-2\epsilon}$ &$\epsilon \in [01, 0.5]$ \\
&  $\mathcal{L}(\theta,x,y_l,y_w)= - \log \sigma \left( \beta \log \frac{\pi_\theta(y_w|x)}{\pi_{\text{ref}}(y_w|x)}  - \beta \log \frac{\pi_\theta(y_l|x)}{\pi_{\text{ref}}(y_l|x)}\right)$ &$\beta \in [0.01, 0.05, 0.1]$ \\ \midrule

CPO~\cite{xu2024contrastive} &  $-\log \sigma  \left(\beta \log \pi_\theta(y_w|x) - \beta \log \pi_\theta(y_l|x) \right) - \lambda \log \pi_\theta (y_w|x)$ & $\lambda = 1.0, \,\, \beta \in [0.01, 0.05, 0.1]$ \\ \midrule
\multirow{2}{*}{KTO \citep{ethayarajh2024kto}} & $-\lambda_w \sigma \left( \beta \log \frac{\pi_\theta(y_w|x)}{\pi_{\text{ref}}(y_w|x)} - z_{\text{ref}} \right) +  \lambda_l \sigma \left( z_{\text{ref}} - \beta \log \frac{\pi_\theta(y_l|x)}{\pi_{\text{ref}}(y_l|x)} \right),\,$ & $\lambda_l = \lambda_w = 1.0$ \\  
& $\text{where} \,\, z_{\text{ref}} = \mathbb{E}_{(x, y) \sim \mathcal{D}} \left[\beta \text{KL}\left( \pi_\theta(y|x) || \pi_{\text{ref}}(y|x) \right)  \right]$ & $\beta \in [0.01, 0.1, 1.0]$ \\ \midrule
\multirow{2}{*}{ORPO \citep{hong2024orpo}} & $-\log p_\theta(y_w|x) - \lambda  \log \sigma \left(\log \frac{p_\theta(y_w|x)}{1 - p_\theta(y_w|x)} - \log \frac{p_\theta(y_l|x)}{1 - p_\theta(y_l|x)}  \right),\,$ & \multirow{2}{*}{$\lambda \in [0.01, 0.1, 1.0]$} \\  
& $\text{where} \,\, p_\theta(y|x) = \exp\left( \frac{1}{|y|} \log \pi_\theta(y|x) \right)$ \\  \midrule
\multirow{2}{*}{R-DPO \citep{park2024disentangling}} & \multirow{2}{*}{$-\log \sigma \left( \beta \log \frac{\pi_\theta(y_w|x)}{\pi_{\text{ref}}(y_w|x)} - \beta \log \frac{\pi_\theta(y_l|x)}{\pi_{\text{ref}}(y_l|x)} + \left(\alpha |y_w| - \alpha |y_l| \right) \right)$} & $\alpha \in [0.05, 0.1, 0.5, 1.0]$ \\
& & $\beta \in [0.01, 0.05, 0.1]$ \\ \midrule
\multirow{2}{*}{SimPO \citep{meng2024simpo}} & \multirow{2}{*}{$-\log \sigma  \left( \frac{\beta}{|y_w|} \log \pi_\theta(y_w|x) - \frac{\beta}{|y_l|} \log \pi_\theta(y_l|x) - \gamma \right)$} & $\beta \in [2.0, 2.5]$ \\
& & $\gamma \in [1.0, 1.2, 1.4, 1.6]$ \\
\bottomrule 
\end{tabular}
}
\caption{Detailed optimization objectives of current preference learning methods. We carefully tune their specific hyperparameters and list the search space in the right column.}
\label{tab:objectives}
\end{table*}

\section{Baseline Methods}
\label{app:baseline}
We compare MPO with other preference optimization methods listed in Table \ref{tab:objectives}. IPO \citep{azar2024general} is a theoretically grounded approach that avoids DPO’s assumption that pairwise preferences can be replaced with pointwise rewards. rDPO \citep{chowdhury2024provably} mitigates the impact of noise on average, making policies trained with this method more robust. CPO \citep{xu2024contrastive} leverages sequence likelihood as a reward and trains jointly with an SFT objective. KTO \citep{ethayarajh2024kto} learns from non-paired preference data, while ORPO \citep{hong2024orpo} introduces a reference-model-free odds ratio term to directly contrast winning and losing responses with the policy model, training it alongside the SFT objective. R-DPO \citep{park2024disentangling} modifies DPO by incorporating an additional regularization term to prevent length exploitation. Finally, SimPO \citep{meng2024simpo} normalizes rewards based on response length and enforces a target reward margin, ensuring that the reward difference between winning and losing responses meets a predefined threshold.

\begin{table}
\centering
\resizebox{\linewidth}{!}{
\begin{tabular}{lccc}
\toprule
& \textbf{Learning Rate} & \textbf{Epoch} & $\beta$ \\
\midrule
LLaMA-3.1-8B-Instruct & 6e-7 &2 &1.0 \\
Gemma-2-9b-it &4e-7 &2 &1.5 \\
Qwen2.5-7B-Instruct &6e-7 &2 &1.5  \\
\bottomrule
\end{tabular}
}
\caption{The hyperparameters in our proposed MPO used for all three backbones.}
\label{tab:hyper}
\end{table}

\section{Implementation Details}
\label{app:implement}

All training experiments are conducted on eight A100 GPUs using the LLaMA-Factory repository \citep{zheng2024llamafactory}. And our MPO is also implement based on this repo. For distributed training, we leverage the DeepSpeed \citep{rasley2020deepspeed} framework with ZeRo-2 optimization. Initially, we perform preliminary experiments to determine optimal batch sizes from [8, 16, 32] and training epochs from [1, 2, 3]. We observe that a batch size of 8 consistently yields the best performance across all methods, while the optimal number of training epochs varies by method. All models on all three backbones are trained with a maximum sequence length of 2048, and we employ a cosine learning rate schedule with a 10\% warmup phase.

To further refine performance, we extensively tune key hyperparameters for all baselines, including the learning rate, training epochs, and method-specific parameters. The learning rate is searched within [3e-7, 4e-7, 5e-7, 6e-7, 1e-6], while training epochs are explored in [1, 2, 3]. Method-specific hyperparameter search spaces are detailed in Table \ref{tab:objectives}. For MPO, $\beta$ is searched in [1.0, 1.5, 2.0] and we find that 1.0 or 1.5 always exhibit the best results across all three backbones. Table \ref{tab:hyper} shows MPO’s hyperparameters used under each backbone. 

\begin{table*}
\centering
\setlength{\extrarowheight}{0pt}
\resizebox{\linewidth}{!}{
\begin{tabular}{lccccccc|cccccccc|c}
\toprule
\textbf{}  & \multicolumn{7}{c|}{\textbf{MultiJail}} & \multicolumn{8}{c|}{\textbf{AdvBench-X}} & \multicolumn{1}{c}{\textbf{CSRT}} \\
& \textbf{En} & \textbf{Zh} & \textbf{Ko} & \textbf{Ar} & \textbf{Bn} & \textbf{Sw} & \textbf{AVG.} & \textbf{En} & \textbf{Zh} & \textbf{Jp} & \textbf{Ko} & \textbf{Ar} & \textbf{Bn} & \textbf{Sw} & \textbf{AVG.} & \textbf{-} \\
\midrule
Qwen2.5 & 12.70	&10.16	&15.87	&15.87	&73.02	&98.10	&37.62 &1.15	&1.35	&5.96	&5.38	&6.35	&57.58	&99.04	&25.26 &34.60\\
\midrule
SFT &10.79	&10.79	&13.02	&\underline{13.02}	&64.76	&99.05	&35.24 &1.35	&2.12	&5.38	&3.46	&5.38	&48.18	&98.46	&23.48 &34.92 \\
DPO &11.43	&10.48	&12.38	&13.33	&69.84	&98.73	&36.03 &1.54	&\underline{1.35}	&5.77	&3.85	&4.81	&51.82	&98.08	&23.89 &37.46 \\
IPO &11.43	&8.89	&13.65	&13.02	&68.25	&99.05	&35.72 &\textbf{0.96}	&1.92	&5.96	&5.19	&5.77	&53.93	&97.70	&24.49 &38.10 \\
rDPO &9.84	&8.25	&15.24	&14.92	&70.16	&98.73	&36.19 &1.73	&1.92	&4.23	&4.04	&6.92	&52.78	&\underline{96.93}	&24.08 &35.56 \\
CPO &13.33	&7.94	&12.38	&13.33	&58.92	&98.73	&34.11 &\underline{1.15}	&2.31	&5.96	&4.23	&6.35	&50.48	&99.04	&24.22 &34.92 \\
KTO &10.16	&9.52	&13.65	&13.97	&66.67	&99.05	&35.50 &2.50	&1.54	&5.38	&4.04	&6.73	&53.93	&98.08	&24.60 &39.81 \\
ORPO &10.16	&10.16	&16.19	&13.97	&67.62	&99.37	&36.25 &2.31 &2.50	&5.77	&3.27	&\underline{4.62}	&48.56	&98.46 &23.64 &30.16 \\
R-DPO &10.79	&\underline{7.62}	&\underline{9.84}	&14.29	&\underline{58.41}	&\underline{98.10}	&\underline{33.18} &1.73	&2.12	&5.58	&4.62	&5.58	&56.24	&98.85	&24.96 &38.41 \\
SimPO &11.75	&9.52	&14.60	&13.97	&70.79	&98.41	&36.51 &1.35	&1.92	&5.38	&4.23	&5.96	&49.33	&98.85	&23.86 &31.43 \\
\midrule
MPO (Ours) & \textbf{7.30}	&\textbf{6.67}	&\textbf{8.89}	&\textbf{13.02} &\textbf{53.65}	&\textbf{92.38}	&\textbf{30.32} &1.92	&\textbf{0.96}	&\textbf{3.27}	&\textbf{2.50}	&\textbf{3.65}	&\textbf{30.33} &\textbf{85.03}	&\textbf{18.24}
 &\textbf{26.35} \\
MPO - En Align &\underline{9.52}	&13.65	&13.33	&13.97	&64.44	&98.73	&35.61 &2.12 &3.46 &\underline{3.27} &\underline{2.88} &5.00 &\underline{46.64} &97.50 &\underline{22.98} &\underline{27.94} \\
\bottomrule
\end{tabular}
}
\caption{Detailed results of Qwen2.5-7B-Instruct on three multilingual safety benchmarks are presented. The evaluation metric used is the Attack Success Rate (ASR), where lower values indicate better performance. The best results achieved by our method and baselines are highlighted in bold, while the second-best results are underlined.}
\label{tab:main_exp_qwen}
\end{table*}

\begin{table}
\centering
\resizebox{\linewidth}{!}{
\begin{tabular}{lcccccccc}
\toprule
\textbf{}  & \multicolumn{8}{c}{\textbf{MT-Bench}} \\
\cmidrule(lr){2-9}
& \textbf{En} & \textbf{Zh} & \textbf{Jp} & \textbf{Ko} & \textbf{Ar} & \textbf{Bn} & \textbf{Sw} & \textbf{AVG.}\\
\midrule
LLaMA-3.1 &7.31	&5.38	&4.88	&5.22	&5.43	&3.98	&3.98	&5.17 \\
\midrule
SFT &7.31 &5.56 &4.84 &4.94 &5.09 &4.25 &3.72 &5.10 \\
DPO &7.44	&5.66	&5.03	&5.49	&4.89	&4.76	&4.16	&5.35 \\
IPO &7.31	&5.42	&4.89  &5.12	&5.23	&4.39	&4.08	&5.26 \\
rDPO &7.31	&5.81	&5.31	&5.16	&5.43	&4.44	&4.21	&5.38 \\
CPO &7.45	&5.59	&4.98	&4.93	&5.04	&4.16	&3.86	&5.14 \\
KTO &7.33	&5.55	&5.02 &5.11	&5.05	&4.39	&4.01	&5.24 \\
ORPO &7.39	&5.41	&4.73	&5.01	&5.36	&4.24	&3.72	&5.12 \\
R-DPO &7.30	&5.63	&5.21	&5.45	&5.48	&4.80	&4.11	&5.43 \\
SimPO &7.48	&5.48	&5.54	&5.21	&5.59	&4.01	&4.11	&5.35 \\
\midrule
MPO & 7.25	&5.32	&5.26	&5.44	&5.38	&4.11	&4.01	&5.25 \\
\midrule
\midrule
Gemma-2 &7.71	&7.07	&6.84	&6.81	&7.06	&5.66	&6.15	&6.76 \\
\midrule
SFT &7.72	&6.86	&6.31	&6.38	&7.08	&5.16	&5.84	&6.48 \\
DPO &7.79	&6.88	&7.06	&6.86	&6.98	&6.26	&6.20	&6.86 \\
IPO &7.61	&6.86	&6.95	&6.86	&7.07	&5.88	&6.17	&6.77 \\
rDPO &7.57	&6.82	&6.93	&6.57	&6.98	&6.03	&6.23	&6.73 \\
CPO &7.73	&6.64	&6.62	&6.56	&7.01	&5.33	&5.98	&6.55 \\
KTO &7.63	&6.87	&7.00	&6.69	&6.88	&6.06	&6.04	&6.74 \\
ORPO &7.71	&6.86	&6.78	&6.49	&7.09	&5.22	&6.07	&6.60 \\
R-DPO &7.77	&7.11	&7.09	&7.33	&6.87	&5.33	&6.11	&6.80 \\
SimPO &7.47	&6.89	&7.00	&6.78	&6.95	&5.84	&6.16	&6.73 \\
\midrule
MPO & 7.83	&6.81	&7.07	&6.88	&7.05	&5.78	&6.16	&6.80 \\
\midrule
\midrule
Qwen2.5 &7.77	&7.36	&6.43	&6.21	&6.60	&4.49	&2.37	&5.89 \\
\midrule
SFT &7.49	&7.14	&6.73	&6.50	&6.46	&4.66	&2.08	&5.87 \\
DPO &7.68	&6.99	&6.79	&6.61	&6.50	&4.71	&2.24	&5.93 \\
IPO &7.66	&7.24	&6.86	&6.29	&6.61	&4.68	&1.98	&5.90 \\
rDPO &7.78	&7.01	&6.58	&6.38	&6.59	&4.48	&2.10	&5.85 \\
CPO &7.61	&7.16	&6.82	&6.33	&6.57	&4.54	&2.12	&5.88 \\
KTO &7.77	&7.03	&6.82	&6.60	&6.64	&4.71	&2.18	&5.96 \\
ORPO &7.52	&7.18	&6.58	&6.43	&6.78	&4.53	&2.09	&5.87 \\
R-DPO &7.61	&7.23	&6.90	&6.33	&6.56	&4.87	&2.20	&5.96 \\
SimPO &7.57	&7.07	&6.73	&6.29	&6.68	&4.77	&2.08	&5.88 \\
\midrule
MPO &7.77	&7.39	&6.71	&6.19	&6.56	&4.43	&2.28	&5.90 \\
\bottomrule 
\end{tabular}
}
\caption{Results on MT-Bench across three backbones.}
\label{tab:mt_bench}
\end{table}

\begin{table}
\centering
\resizebox{\linewidth}{!}{
\begin{tabular}{lcccccccc}
\toprule
\textbf{}  & \multicolumn{8}{c}{\textbf{M-MMLU}} \\
\cmidrule(lr){2-9}
& \textbf{En} & \textbf{Zh} & \textbf{Jp} & \textbf{Ko} & \textbf{Ar} & \textbf{Bn} & \textbf{Sw} & \textbf{AVG.}\\
\midrule
LLaMA-3.1 &67.70 &51.30	&47.90	&43.30	&47.60	&41.40	&40.60	&48.54 \\
\midrule
SFT &67.30	&50.90	&47.70	&47.00	&42.60	&40.60	&39.20	&47.90 \\
DPO &67.10	&51.50	&48.40	&47.30	&41.80	&39.80	&38.20	&47.73 \\
IPO &67.30	&51.60	&48.30	&47.90	&43.00	&41.30	&39.60	&48.43 \\
rDPO &67.20	&50.80	&47.90	&47.30	&42.80	&40.00	&40.50	&48.07 \\
CPO &67.40	&51.60	&47.60	&47.50	&42.50	&40.00	&39.80	&48.06 \\
KTO &67.10	&51.40	&48.50	&47.60	&41.90	&40.40	&40.40	&48.19 \\
ORPO &67.30	&51.00	&48.10	&47.30	&42.10	&41.30	&38.90	&48.00 \\
R-DPO &66.90 &51.50	&48.10	&47.70	&43.30	&40.70	&40.20	&48.34 \\
SimPO &66.70 &51.40	&47.70	&47.80	&43.00	&40.90	&40.40	&48.27 \\
\midrule
MPO &67.10 &50.70	&48.40	&42.40	&47.70	&40.10	&38.70	&47.87 \\
\midrule
\midrule
Gemma-2 &73.40	&61.20	&59.40	&53.80	&59.10	&49.90	&52.40	&58.45 \\
\midrule
SFT &61.10	&50.70	&73.30	&59.40	&59.40	&55.40	&52.50	&58.83 \\
DPO &61.20	&49.80	&73.40	&58.70	&59.20	&53.90	&52.20	&58.34 \\
IPO &61.30	&49.70	&73.30	&59.30	&59.50	&53.60	&52.40	&58.44 \\
rDPO &61.00	&50.20	&73.30	&59.10	&59.40	&54.20	&52.50	&58.53 \\
CPO &61.40	&50.80	&73.30	&59.30	&59.60	&55.10	&52.60	&58.87 \\
KTO &61.40	&49.90	&73.40	&59.20	&59.50	&53.90	&52.40	&58.53 \\
ORPO &61.70	&50.40	&73.20	&59.30	&59.70	&55.20	&52.90	&58.91 \\
R-DPO &60.80	&49.00	&73.30	&59.10	&59.20	&54.00	&51.40	&58.11 \\
SimPO &61.30	&49.90	&73.30	&59.20	&59.40	&53.90	&52.30	&58.47 \\
\midrule
MPO &73.40 &61.20	&58.90	&54.00	&59.50	&49.80	&52.10	&58.41 \\
\midrule
\midrule
Qwen2.5 &72.50	&63.90	&57.70	&49.70	&56.60	&43.20	&31.50	&53.59 \\
\midrule
SFT &64.10	&43.10	&72.60	&56.70	&57.80	&49.30	&31.30	&53.56 \\
DPO &64.00	&43.10	&72.50	&56.70	&57.70	&49.50	&31.40	&53.56 \\
IPO &64.20	&43.10	&72.20	&57.00	&57.30	&49.70	&31.40	&53.56 \\
rDPO &64.00	&43.40	&72.40	&56.60	&57.70	&49.70	&31.40	&53.60 \\
CPO &64.40	&42.60	&72.70	&57.00	&57.90	&49.10	&31.60	&53.61 \\
KTO &64.00	&43.20	&72.50	&56.70	&57.70	&49.80	&31.40	&53.61 \\
ORPO &64.50	&43.00	&72.70	&57.10	&57.80	&50.40	&31.50	&53.86 \\
R-DPO &64.20	&43.30	&72.50	&56.50	&57.60	&49.60	&31.40	&53.59 \\
SimPO &64.10	&43.10	&72.20	&56.80	&57.70	&50.00	&31.40	&53.61 \\
\midrule
MPO &72.20	&64.50	&57.30	&50.50	&57.10	&43.20	&31.60	&53.77 \\
\bottomrule 
\end{tabular}
}
\caption{Results on M-MMLU across three backbones.}
\label{tab:mmlu}
\end{table}

\begin{table}
\centering
\resizebox{\linewidth}{!}{
\begin{tabular}{lccccc}
\toprule
\textbf{}  & \multicolumn{5}{c}{\textbf{MGSM}} \\
\cmidrule(lr){2-6}
& \textbf{En} & \textbf{Zh} & \textbf{Bn} & \textbf{Sw} & \textbf{AVG.}\\
\midrule
LLaMA-3.1 &88.00 &67.20	&12.40	&40.80	&52.10 \\
\midrule
SFT &86.80	&69.60	&13.60	&54.00	&56.00 \\
DPO &85.60	&68.80	&15.60	&45.20	&53.80 \\
IPO &85.60	&68.80	&16.00	&46.80	&54.30 \\
rDPO &86.80	&71.20	&13.60	&39.60	&52.80 \\
CPO &88.80	&73.60	&11.20	&43.20	&54.20 \\
KTO &84.80	&67.60	&16.80	&48.40	&54.40 \\
ORPO &86.80	&69.60	&14.80	&55.20	&56.60 \\
R-DPO &86.80	&71.20	&13.60	&39.60	&52.80 \\
SimPO &86.00	&72.40	&11.60	&46.40	&54.10 \\
\midrule
MPO &88.00 &68.40	&12.00	&53.60	&55.50 \\
\midrule
\midrule
Gemma-2 &90.00	&77.60	&66.00	&75.20	&77.20 \\
\midrule
SFT &89.60	&79.20	&43.60	&63.60	&69.00 \\
DPO &89.20	&78.80	&67.60	&73.60	&77.30 \\
IPO &90.00	&78.40	&67.20	&75.20	&77.70 \\
rDPO &90.40	&78.00	&65.60	&76.80	&77.70 \\
CPO &90.40	&79.20	&46.00	&68.00	&70.90 \\
KTO &90.00	&77.60	&67.60	&75.20	&77.60 \\
ORPO &90.40	&80.00	&49.60	&65.20	&71.30 \\
R-DPO &90.00	&77.60	&75.20	&75.60	&79.60 \\
SimPO &90.00	&76.80	&67.20	&75.20	&77.30 \\
\midrule
MPO &90.80	&80.40	&70.00	&74.00	&78.80 \\
\midrule
\midrule
Qwen2.5 &87.20	&82.00	&35.20	&6.40	&52.70 \\
\midrule
SFT &87.60	&82.00	&35.60	&8.40	&53.40 \\
DPO &87.60	&82.00	&38.00	&7.20	&53.70 \\
IPO &88.00	&84.40	&36.40	&7.60	&54.10 \\
rDPO &88.00	&82.40	&36.00	&8.00	&53.60 \\
CPO &88.80	&82.40	&34.00	&9.20	&53.60 \\
KTO &87.60	&82.40	&36.40	&7.60	&53.50 \\
ORPO &88.40	&82.40	&24.00	&6.80	&50.40 \\
R-DPO &87.20	&82.40	&36.40	&7.20	&53.30 \\
SimPO &88.40	&84.00	&38.00	&9.20	&54.90 \\
\midrule
MPO &88.40	&82.80 &34.40	&6.80	&53.10 \\
\bottomrule 
\end{tabular}
}
\caption{Results on MGSM across three backbones.}
\label{tab:mgsm}
\end{table}

\section{Additional Experimental Results}

\subsection{Results on Qwen2.5}
\label{app:main_exp_qwen}
Table \ref{tab:main_exp_qwen} demonstrates the performance comparison of MPO and baselines based on Qwen2.5-7B-Instruct. we have drawn the following key insights:

\paragraph{MPO still exhibits robust and consistent performance across various benchmarks and maintain multilingual utility.} It consistently surpasses all preference learning methods, highlighting its outstanding safety alignment capabilities and scalability. Tables \ref{tab:mt_bench}, \ref{tab:mmlu} and \ref{tab:mgsm} presents MPO maintains multilingual utility on MT-Bench, M-MMLU and MGSM, respectively.

\paragraph{Multilingual safety alignment depends on foundational abilities} The improvement of multilingual safety performance relies on the foundational multilingual capabilities of the backbone model. Results on Qwen2.5 show that while MPO still achieves significant gains compared to the original model and baselines, its absolute performance lags behind other two backbones, especially for low-resource languages. This disparity arises from Qwen2.5’s weaker foundational abilities in these languages. More Specifically, as shown in Table \ref{tab:mt_bench}, Qwen2.5 exhibits weak instruction-following ability in Bn and Sw, frequently generating outputs unrelated to the input. In our evaluation, such outputs are classified as unsafe.

\subsection{Evaluation on Multilingual Utility}
\label{app:utility}

\paragraph{Evaluation Settings}
We conduct a comprehensive evaluation of MPO’s impact on multilingual utility across the following benchmarks.

\begin{itemize}
    \item \textbf{MT-Bench} \citep{zheng2023judging}: The dataset is designed for open-ended generation to evaluate a model’s ability to follow multi-turn instructions. In our experimental setup, this benchmark covers English (En), Chinese (Zh), Arabic (Ar), Japanese (Jp), Korean (Ko), Swahili (Sw) and Bengali (Bn). We collect data in English\footnote{\url{https://huggingface.co/datasets/HuggingFaceH4/mt_bench_prompts}}, Japanese\footnote{\url{https://huggingface.co/datasets/shi3z/MTbenchJapanese}}, Korean\footnote{\url{https://huggingface.co/datasets/StudentLLM/Korean_MT-Bench_questions}}, and Arabic\footnote{\url{https://huggingface.co/spaces/QCRI/mt-bench-ar/tree/main/data/mt_bench_ar}} from huggingface, and Chinese\footnote{\url{https://github.com/HIT-SCIR/huozi}} from github. In addition, we use GPT-4o to translate the English data into Swahili and Bengali, and performed manual proofreading to ensure correctness. The evaluation follows the \textbf{LLM-as-a-judge} approach, where GPT-4o is prompted to assign a score directly to a single response on a scale of 1 to 10. It is essential to highlight that the languages targeted for enhancement, as mentioned above, are all within the capability range of GPT-4o, especially given that its official model card \citep{openai2024gpt4o} emphasizes support for low-resource languages such as Swahili (Sw) and Bengali (Bn). This underscores the validity and reliability of the evaluation approach.
    \item \textbf{M-MMLU} \citep{hendrycks2021measuring}:\footnote{\url{https://huggingface.co/datasets/openai/MMMLU}} The MMLU is a widely recognized benchmark of general knowledge attained by AI models. It covers a broad range of topics from 57 different categories, covering elementary-level knowledge up to advanced professional subjects like law, physics, history, and computer science. OpenAI translated the MMLU’s test set into 14 languages using professional human translators. Relying on human translators for this evaluation increases confidence in the accuracy of the translations, especially for low-resource languages. In our experimental setup, we adopt the 5-shot evaluation and this benchmark covers English (En), Chinese (Zh), Japanese (Jp), Arabic (Ar), Korean (Ko), Swahili (Sw) and Bengali (Bn).
    \item \textbf{MGSM} \citep{shi2023language}:\footnote{\url{https://huggingface.co/datasets/juletxara/mgsm}} Multilingual Grade School Math Benchmark (MGSM) is a benchmark of grade-school math problems. The same 250 problems from GSM8K \citep{cobbe2021training} are each translated via human annotators in 10 languages. The dataset was created to support the task of question answering on basic mathematical problems that require multi-step reasoning. In our experimental setup, we performance evaluation via 0-shot CoT \citep{wei2022chain} and this benchmark covers English (En), Chinese (Zh), Swahili (Sw) and Bengali (Bn).
\end{itemize}

\paragraph{Detailed Results} Here, we demonstrate the detailed results and the comparison with baseline methods. For MT-Bench, results on LLaMA-3.1-8B-Instruct, Gemma-2-9B-it and Qwen2.5-7B-Instruct are shown in Table \ref{tab:mt_bench}. For M-MMLU, results are shown in Table \ref{tab:mmlu}. For MGSM, results are shown in Table \ref{tab:mgsm}. The results across three backbones show that MPO consistently maintains the general utility of both the dominant and target languages, demonstrating its effectiveness in achieving multilingual safety alignment without compromising the model’s multilingual utility.

Current preference learning methods typically incorporate a KL constraint during training to prevent the model from deviating too far from its original state, ensuring that multilingual general utility is well preserved. As a result, these methods maintain multilingual capabilities comparable to the original model, even after alignment.

Under the same achievement of preserving multilingual general utility, MPO achieves significantly superior multilingual safety performance compared to these methods. By leveraging reward gap minimization with the dominant language as a high-quality supervision signal, MPO effectively transfers safety alignment across languages without degrading the model’s overall linguistic competence. This highlights its advantage in balancing multilingual safety and utility, making it a more effective approach for multilingual safety alignment in LLMs. And the comparison of KL constraint and the representation constraint used in MPO is further discussed in Appendix \ref{app:ablation}.

\begin{table*}
\centering
\setlength{\extrarowheight}{0pt}
\resizebox{\linewidth}{!}{
\begin{tabular}{lccccccc|cccccccc}
\toprule
\textbf{}  & \multicolumn{7}{c|}{\textbf{MultiJail}} & \multicolumn{8}{c}{\textbf{MT-Bench}} \\
& \textbf{En} & \textbf{Zh} & \textbf{Ko} & \textbf{Ar} & \textbf{Bn} & \textbf{Sw} & \textbf{AVG.} & \textbf{En} & \textbf{Zh} & \textbf{Jp} & \textbf{Ko} & \textbf{Ar} & \textbf{Bn} & \textbf{Sw} & \textbf{AVG.} \\
\midrule
LLaMA-3.1 & 14.60 & 20.32 & 52.38 & 16.83 & 49.52 & 37.78 & 31.91 &7.31	&5.38	&4.88	&5.22	&5.43	&3.98	&3.98	&5.17\\
\midrule
Constant 0.1 &25.08	&46.67	&61.90	&63.38	&59.37	&85.08	&56.91 &7.13	&5.18	&4.72	&4.47	&4.67	&4.16	&3.39	&4.82 \\
Constant 0.5 &10.16	&8.89	&26.98	&14.29	&21.59	&45.08	&21.17	&\textbf{7.34}	&\textbf{5.49}	&4.77	&5.23	&5.04	&4.08	&3.68	&5.09 \\
Constant 1.0 &2.86	&1.59	&11.11	&1.90	&7.30	&18.41	&7.20	&7.12	&5.26	&4.71	&5.09	&5.08	&3.97	&3.71	&4.99 \\
Constant 1.58 &2.86	&1.27	&6.03	&0.32	&6.35	&13.97	&5.13	&7.26	&5.16	&5.16	&5.18	&5.02	&\textbf{4.27}	&\textbf{4.14}	&5.17 \\
Constant 2.0 &\textbf{0.32}	&\textbf{0.00}	&\textbf{0.63}	&\textbf{0.00}	&\textbf{5.71}	&6.03	&\textbf{2.12}	&7.04	&4.43	&4.01	&5.26	&3.43	&3.61	&3.54	&4.47 \\
Constant 5.0 &\textbf{0.32}	&\textbf{0.00}	&\textbf{0.63}	&1.90	&7.30	&9.21	&3.23	&\textbf{7.34}	&4.81	&4.53	&5.22	&3.88	&4.00	&3.02	&4.69 \\
Constant 10.0 &\textbf{0.32}	&0.32	&0.95	&0.63	&6.35	&6.03	&2.43	&6.88	&4.16	&3.64	&4.20	&3.06	&3.63	&3.58	&4.16 \\
Constant 20.0 &\textbf{0.32}	&0.32	&\textbf{0.63}	&0.63	&14.60	&\textbf{3.81}	&3.39	&7.21	&4.94	&4.31	&5.16	&4.03	&3.26	&3.61	&4.65 \\
\midrule
MPO (Ours) & 2.22	&0.95	&4.76	&1.90	&12.38	&10.79	&5.98 & 7.25	&5.32	&\textbf{5.26}	&\textbf{5.44}	&\textbf{5.38}	&4.11	&4.01	&\textbf{5.25} \\
\bottomrule
\end{tabular}
}
\caption{Results of the multilingual safety performance and general utility of the model when replacing the dominant language reward gap with a fixed value ranging from 0.1 to 20. The best results are highlighted in bold.}
\label{tab:fix_value}
\end{table*}

\begin{table}
\centering
\resizebox{\linewidth}{!}{
\begin{tabular}{lccccccc}
\toprule
\textbf{}  & \multicolumn{7}{c}{\textbf{MultiJail}} \\
\cmidrule(lr){2-8}
& \textbf{En} & \textbf{Zh} & \textbf{Ko} & \textbf{Ar} & \textbf{Bn} & \textbf{Sw} & \textbf{AVG.}\\
\midrule
LLaMA-3.1 & 14.60 & 20.32 & 52.38 & 16.83 & 49.52 & 37.78 & 31.91 \\
\midrule
SDRRL &4.76	&4.13	&18.41	&3.81	&21.90	&21.59	&12.43 \\
CLA &11.75	&15.24	&43.49	&12.87	&42.54	&55.56	&30.24 \\
\textsc{Lens} &16.09	&60.95	&55.24	&25.08	&60.95	&80.32	&49.77 \\
\midrule
MPO &\textbf{2.22}	&\textbf{0.95}	&\textbf{4.76}	&\textbf{1.90}	&\textbf{12.38}	&\textbf{10.79}	&\textbf{5.98} \\
\bottomrule
\end{tabular}
}
\caption{Comparison with cross-lingual transfer method on MultiJail. The evaluation metric used is the Attack Success Rate (ASR), where lower values indicate better performance. The best results are highlighted in bold.}
\label{tab:cross_lingual}
\end{table}

\begin{table*}
\centering
\setlength{\extrarowheight}{0pt}
\resizebox{\linewidth}{!}{
\begin{tabular}{lccccccc|cccccccc}
\toprule
\textbf{}  & \multicolumn{7}{c|}{\textbf{MultiJail}} & \multicolumn{8}{c}{\textbf{MT-Bench}} \\
& \textbf{En} & \textbf{Zh} & \textbf{Ko} & \textbf{Ar} & \textbf{Bn} & \textbf{Sw} & \textbf{AVG.} & \textbf{En} & \textbf{Zh} & \textbf{Jp} & \textbf{Ko} & \textbf{Ar} & \textbf{Bn} & \textbf{Sw} & \textbf{AVG.} \\
\midrule
LLaMA-3.1 & 14.60 & 20.32 & 52.38 & 16.83 & 49.52 & 37.78 & 31.91 &7.31	&5.38	&4.88	&5.22	&5.43	&3.98	&3.98	&5.17\\
\midrule
MPO	&\textbf{2.22}	&\textbf{0.95}	&\textbf{4.76}	&\textbf{1.90}	&\textbf{12.38}	&\textbf{10.79}	&\textbf{5.98}	&\textbf{7.25}	&\textbf{5.32}	&\textbf{5.26}	&\textbf{5.44}	&\textbf{5.38}	&4.11	&\textbf{4.01}	&\textbf{5.25} \\
MPO - Policy &56.51	&48.89	&70.79	&46.35	&81.27	&84.13	&64.66	&7.04	&5.23	&5.03	&5.34	&5.36	&4.11	&3.97	&5.15 \\
\midrule
\midrule
Gemma-2	&2.54	&9.52	&14.61	&4.13	&20.32	&14.60	&10.95	&7.71	&7.07	&6.84	&6.81	&7.06	&5.66	&6.15	&6.76 \\
\midrule
MPO	&\textbf{0.63}	&\textbf{4.76}	&\textbf{6.98}	&\textbf{3.81}	&\textbf{16.51}	&\textbf{7.94}	&\textbf{6.77}	&\textbf{7.83}	&\textbf{6.81}	&\textbf{7.07}	&\textbf{6.88}	&\textbf{7.05}	&\textbf{5.78}	&\textbf{6.16}	&\textbf{6.80} \\
MPO - Policy	&1.59	&5.08	&7.30	&5.40	&17.78	&10.83	&7.99	&7.81	&6.77	&6.73	&6.59	&6.84	&5.71	&6.01	&6.64 \\
\midrule
\midrule
Qwen-2.5	&12.70	&10.16	&15.87	&15.87	&73.02	&98.10	&37.62	&7.77	&7.36	&6.43	&6.21	&6.60	&4.49	&2.37	&5.89 \\
\midrule
MPO	&\textbf{7.30}	&\textbf{6.67}	&\textbf{8.89}	&\textbf{13.02}	&\textbf{53.65}	&\textbf{92.38}	&\textbf{30.32}	&\textbf{7.77}	&\textbf{7.39}	&\textbf{6.71}	&\textbf{6.19}	&\textbf{6.56}	&\textbf{4.43}	&\textbf{2.28}	&\textbf{5.90} \\
MPO - Policy	&14.60	&10.48	&12.70	&13.33	&60.63	&98.73	&35.08	&7.69	&6.83	&6.57	&6.16	&6.48	&4.13	&2.17	&5.72 \\
\bottomrule
\end{tabular}
}
\caption{Results of the multilingual safety performance and general utility of the model when replacing the computation of the dominant language reward gap with the policy model itself.}
\label{tab:policy_reward}
\end{table*}

\subsection{Ablation Study}
\label{app:ablation}

\paragraph{Fixed Constants as the Supervision Signal} Table \ref{tab:fix_value} presents the detailed results of multilingual safety performance and general utility of the model when replacing the dominant language reward gap with a fixed value ranging from 0.1 to 20. Notably, 1.58 corresponds to the average reward gap of dominant language samples in the training set. As the constant increases, multilingual safety performance steadily improves, even exceeding the performance of models aligned using the actual dominant language reward gap. However, this improvement comes at a significant cost to multilingual general utility, as excessive alignment strength induces substantial parameter shifts, leading to model collapse despite the application of a retention constraint. Additionally, setting the constant to 1.58 yields only limited improvements, suggesting that fine-grained supervision at the sample level is superior to coarse-grained dataset-level alignment.

\paragraph{Reward Gap of Other Languages as the Supervision Signal}
Table \ref{tab:other_lang} further demonstrates that using the reward gap of a target language as the alignment objective fails to yield meaningful safety improvements. When selecting the second-best safety-performing language (Arabic) or even low-resource languages (Swahili, Bengali), no effective multilingual safety enhancement is observed. This reinforces that the dominant language’s reward gap provides a more reliable and high-quality alignment supervision signal.

\paragraph{Comparison with Cross-Lingual Transfer Methods} Cross-lingual transfer methods posit that skills acquired in one source language can be effectively transferred to other languages \citep{huang2023not,ranaldi2023empowering,qin2023cross,etxaniz2024multilingual}. This has been achieved through two main approaches: aligning multilingual representations with the activation space of LLMs: CLA \citep{li2024improving} and \textsc{Lens} \citep{zhao2024lens}, or distilling knowledge from the dominant language: SDRRL \citep{zhang2024enhancing}. The details of these recent advancements are as follows:
\begin{itemize}
    \item \textbf{CLA}: It aligns internal sentence representations across languages through multilingual contrastive learning and ensures output alignment by adhering to cross-lingual instructions in the target language.
    \item \textbf{\textsc{Lens}}: It enhances multilingual capabilities by leveraging LLMs’ internal language representation spaces. \textsc{Lens} operates on two subspaces: the language-agnostic subspace, where it aligns target languages with the central language to inherit strong semantic representations, and the language-specific subspace, where it separates target and central languages to preserve linguistic specificity.
    \item \textbf{SDRRL}: It leverages self-distillation from resource-rich languages to effectively enhance multilingual performance through the use of self-distilled data.
\end{itemize}
Table \ref{tab:cross_lingual} demonstrates that MPO consistently outperforms these methods, maintaining strong multilingual safety alignment. This further highlights the advantage of leveraging the dominant language’s reward gap as a fine-grained supervision signal. Unlike MPO, which explicitly minimizes the reward gap difference between the dominant language and target languages, existing cross-lingual transfer approaches struggle with noisy preference signals and suboptimal knowledge transfer. Additionally, they often exhibit performance degradation in low-resource languages, where data scarcity amplifies alignment instability.

\begin{table}
\centering
\resizebox{\linewidth}{!}{
\begin{tabular}{lccccccc}
\toprule
\textbf{}  & \multicolumn{7}{c}{\textbf{MultiJail}} \\
\cmidrule(lr){2-8}
& \textbf{En} & \textbf{Zh} & \textbf{Ko} & \textbf{Ar} & \textbf{Bn} & \textbf{Sw} & \textbf{AVG.}\\
\midrule
MAPO - DPO &5.40	&3.49	&15.87	&3.17	&27.71	&42.86	&16.42 \\
MAPO - MPO &\textbf{2.46}	&\textbf{1.59}	&\textbf{3.17}	&\textbf{3.17}	&\textbf{9.52}	&\textbf{8.25}	&\textbf{4.69} \\
\midrule
LIDR - DPO + NLL &\textbf{1.90}	&6.03	&18.10	&19.37	&80.00	&81.27	&34.45 \\
LIDR - MPO &2.22	&\textbf{2.86}	&\textbf{1.90}	&\textbf{6.67}	&\textbf{8.89}	&\textbf{11.12}	&\textbf{5.61} \\
\bottomrule
\end{tabular}
}
\caption{Results of the multilingual safety performance on MultiJail. The evaluation metric used is the Attack Success Rate (ASR), where lower values indicate better performance. The best results are highlighted in bold.}
\label{tab:data_source}
\end{table}

\paragraph{Effect of Constant Reward Gap from the Domiant Language} We consider an alternative design of MPO---namely, computing the dominant language reward gap using the policy model instead of the reference model (MPO-Policy).

We implement MPO - Policy on all three backbones, LLaMA-3.1, Gemma-2 and Qwen2.5, where the dominant reward gap is computed using the policy model during training. We performed an equivalent hyperparameter search over learning rates [3e-7, 4e-7, 5e-7, 6e-7], training epochs [1, 2, 3], and  values [1.0, 1.5, 2.0], and report the best-performing configuration. The results, shown in the Table \ref{tab:policy_reward}, indicate that MPO - Policy consistently underperforms compared to our proposed method across all three backbones, both in terms of safety alignment (as measured by MultiJail, the lower score is better) and general capabilities (as measured by MT-Bench, the higher score is better). This empirically supports our choice to use the reference-based reward gap.

We further provide theoretical justification. When computing $\mathrm{RG}^d$ using the current model $\pi_\theta$ instead of the reference model $\pi_{ref}$, The gradient of this loss function becomes:
\begin{equation}\small
\label{eq:grad_2}
\begin{split}
\nabla_\theta& \mathcal{L}1 = 2\beta \ \mathbb{E}\mathcal{D} \Big[(\beta \cdot \mathrm{RG}^t - \mathrm{RG}^d) \cdot \\ \Big(
& \left( \frac{1}{|y_w^t|} \nabla_\theta \log \pi_\theta(y_w^t|x^t) - \frac{1}{|y_l^t|} \nabla_\theta \log \pi_\theta(y_l^t|x^t) \right) - \\
& \left( \frac{1}{|y_w^d|} \nabla_\theta \log \pi_\theta(y_w^d|x^d) - \frac{1}{|y_l^d|} \nabla_\theta \log \pi_\theta(y_l^d|x^d) \right)
\Big) \Big]
\end{split}
\end{equation}
By comparing the gradients of the two formulations, Equation \ref{eq:grad_1} and Equation \ref{eq:grad_2}, obtaining dominant reward gap from the policy model  has the following drawbacks:
\begin{itemize}
    \item \textbf{Optimization instability}: the direction of the gradient can fluctuate significantly as both sides of the loss are functions of $\theta$.
    \item \textbf{Lack of anchoring}: without a stable reference, the loss can converge to trivial solutions where both $\mathrm{RG}^t$ and $\mathrm{RG}^d$ collapse toward zero, rather than aligning their structure.
\end{itemize}
In summary, our current design using reward gap from reference model not only improves training stability but also provides a clearer learning signal, enabling more reliable cross-lingual safety alignment. This design choice is well-justified, as the dominant language in the original model typically exhibits the strongest safety alignment

\subsection{Impact of Data Source}
\label{app:data_source}
Recent studies explore the use of LLMs themselves to generate multilingual preference data, rather than relying on external translation tools. Two notable approaches in this direction are MAPO \citep{she2024mapo} and LIDR \citep{yang2024language}.
\begin{itemize}
    \item MAPO constructs multilingual preference data by sampling multiple responses from an LLM in a given target languages and ranking them based on an alignment score computed via an external translation model, which measures their consistency with the response in the dominant language. The ranked responses form preference pairs that are then optimized using DPO \citep{rafailov2023direct}.
    \item LIDR relies on the LLM’s own translation capability to convert English preference data into target languages, followed by DPO optimization with an additional NLL loss.
\end{itemize}

While both methods explore multilingual preference data generation, they do not propose improvements to the multilingual preference optimization process itself, instead relying solely on DPO. 

To evaluate whether MPO remains effective when trained on preference data obtained using these methods, we conduct experiments using MAPO- and LIDR-generated data. As shown in Table \ref{tab:data_source}, MPO consistently achieves the best multilingual safety alignment results across both data sources, demonstrating its robustness to variations in preference data. These results emphasize that while MAPO and LIDR explore multilingual preference data construction, they do not address the fundamental challenges of multilingual preference optimization. MPO, in contrast, not only adapts to different multilingual data sources but also improves the optimization process, ensuring stable and effective multilingual safety alignment.

\begin{figure*}
    \centering
    \subfigure[The visualization of multilingual representations for English and Chinese on LLaMA-3.1-8B-Instruct.]{\includegraphics[width=0.32\textwidth]{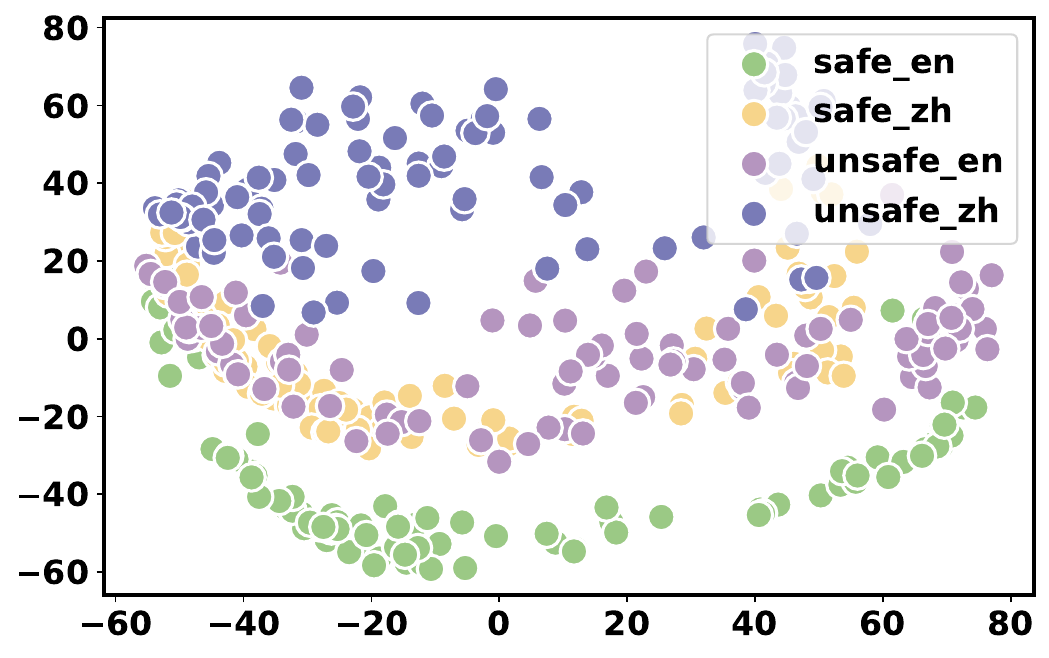}}
    \subfigure[The visualization of multilingual representations for English and Arabic on LLaMA-3.1-8B-Instruct.]{\includegraphics[width=0.32\textwidth]{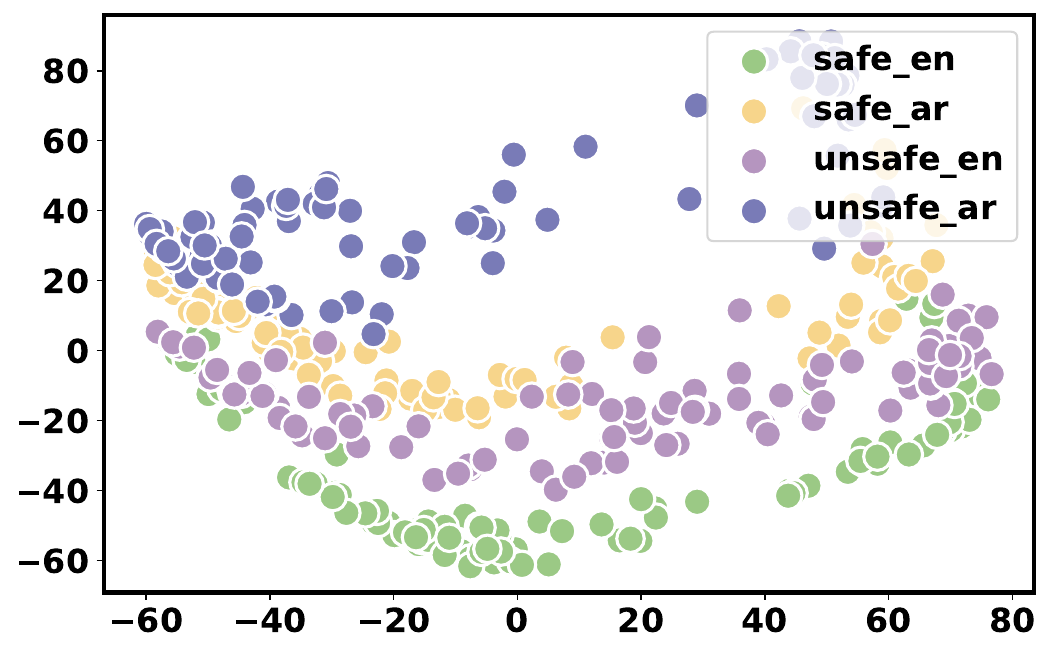}}
    \subfigure[The visualization of multilingual representations for English and Swahili on LLaMA-3.1-8B-Instruct.]{\includegraphics[width=0.32\textwidth]{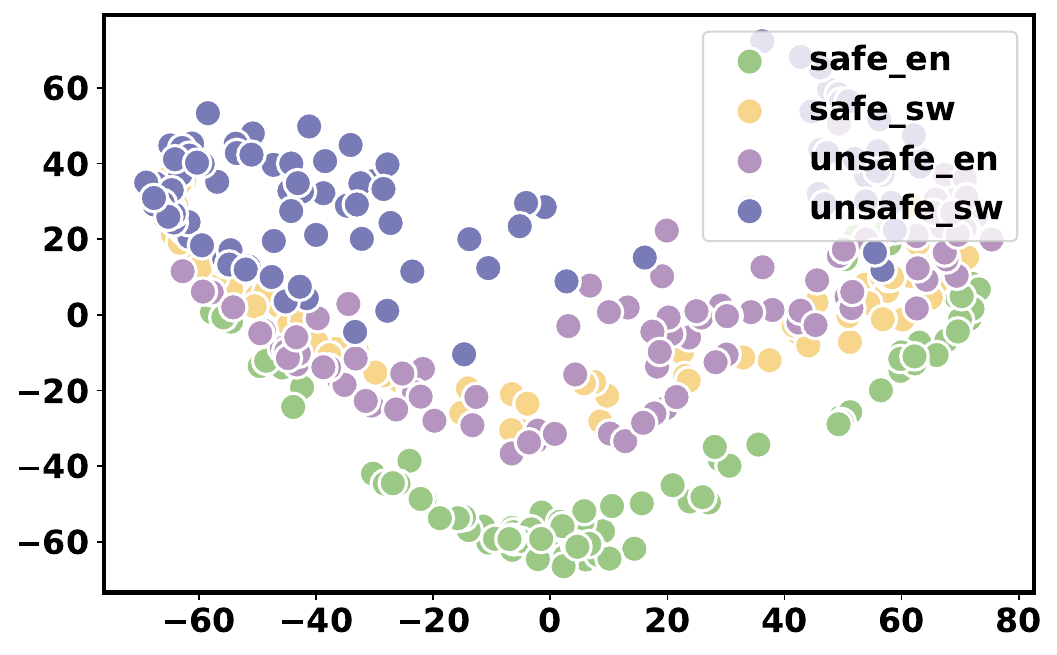}}\\
    \subfigure[The visualization of multilingual representations for English and Chinese on LLaMA-3.1-8B-Instruct after multilingual safety alignment via MPO.]{\includegraphics[width=0.32\textwidth]{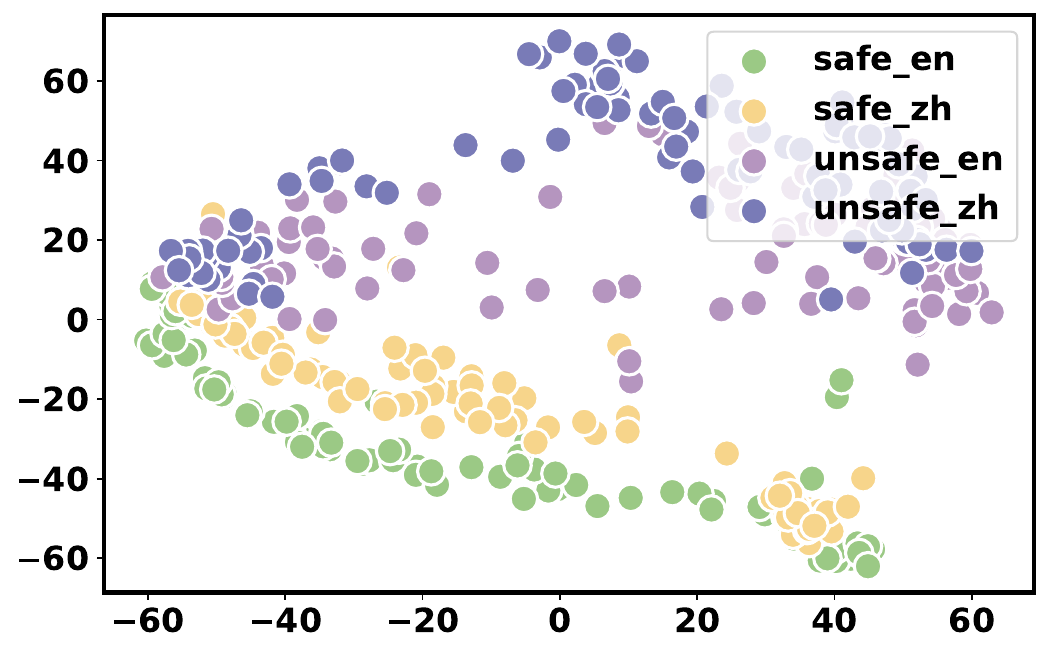}}
    \subfigure[The visualization of multilingual representations for English and Arabic on LLaMA-3.1-8B-Instruct after multilingual safety alignment via MPO.]{\includegraphics[width=0.32\textwidth]{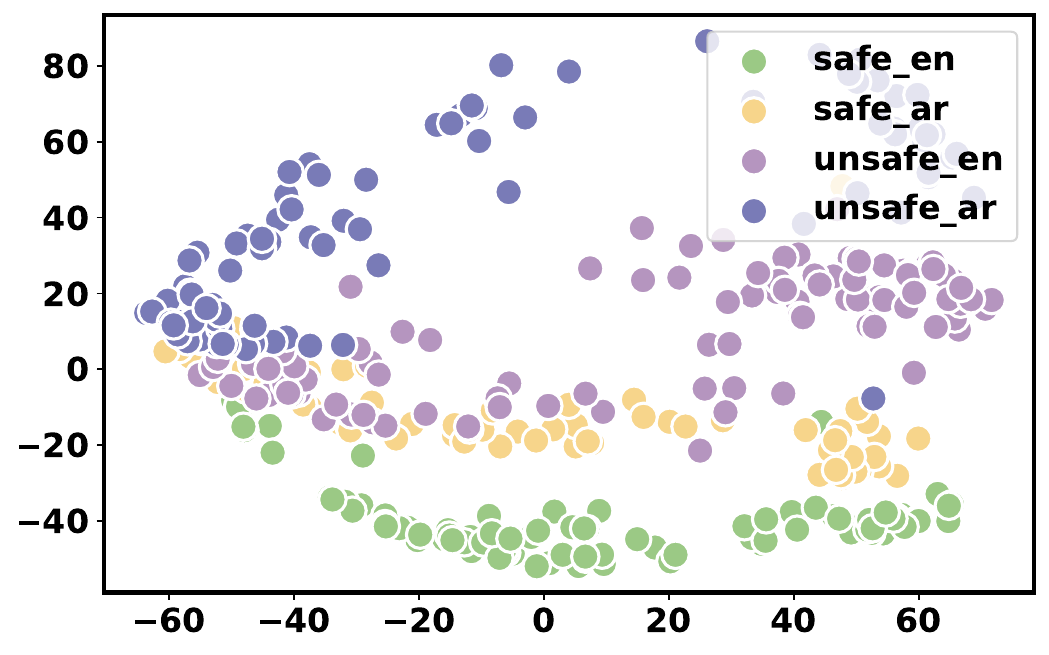}}
    \subfigure[The visualization of multilingual representations for English and Swahili on LLaMA-3.1-8B-Instruct after multilingual safety alignment via MPO.]{\includegraphics[width=0.32\textwidth]{figs/visual_MPO_en_sw.pdf}}
    \caption{The visualization of multilingual representations for safe and unsafe inputs across different languages. The upper row (a-c) illustrates the representation space of the original LLaMA-3.1-8B-Instruct model for English (En) and three additional languages: Chinese (Zh), Arabic (Ar), and Swahili (Sw). The lower row (d-f) presents the corresponding representation space after applying multilingual safety alignment via MPO. The visualizations highlight the structural changes in the representation space induced by MPO alignment.}
    \label{fig:visual_all}
\end{figure*}

\subsection{Visualization Analysis}
\label{app:visual}
To further understand what MPO brings for the multilingual safety alignment of LLMs, as shown in Figure \ref{fig:visual_all}, we perform Principal Component Analysis (PCA) to visualize the multilingual representations in the activation spaces. Specifically, the multilingual harmful inputs are sourced from the AdvBench-X dataset. For each input, we append both a corresponding safe response and an unsafe response to visualize the model’s representation. All representations are extracted from the final layer of the model’s output and the backbone model is LLaMA-3.1-8B-Instruct. 

Notably, in all cases, the boundary between safe and unsafe inputs in English remains consistently clear. However, in the original model (a-c), the distinction between safe and unsafe inputs in the target languages (Zh, Ar, Sw) appears less structured and more entangled. After applying MPO alignment (d-f), the model demonstrates a significantly improved separation of safe and unsafe inputs in the target languages. This indicates that MPO enhances the multilingual safety alignment of the model, allowing it to develop clearer decision boundaries in languages beyond English.

\begin{table}
\centering
\resizebox{\linewidth}{!}{
\begin{tabular}{lcccccccc}
\toprule
\textbf{}  & \multicolumn{8}{c}{\textbf{X-AdvBench}} \\
\cmidrule(lr){2-9}
& \textbf{En} & \textbf{Zh} & \textbf{Jp} & \textbf{Ko} & \textbf{Ar} & \textbf{Bn} & \textbf{Sw} & \textbf{AVG.}\\
\midrule
Aya-101 &17.88	&20.00	&42.69	&45.38	&20.32	&58.08	&53.93	&36.90 \\
+ MPO  &\textbf{8.65} &\textbf{14.04} &\textbf{20.19} &\textbf{25.19} &\textbf{9.81} &\textbf{31.35}	&\textbf{39.23}	&\textbf{21.21} \\
\bottomrule
\end{tabular}
}
\caption{Results of the multilingual safety performance on X-AdvBench. The backbone is Aya-101. The best results are highlighted in bold.}
\label{tab:aya_result}
\end{table}

\subsection{Results on Aya-101}

We includes results on an explicitly multilingual model Aya-101, providing more valuable empirical insights. The results are summarized in Table \ref{tab:aya_result}.

Although Aya-101 supports a wide range of languages, its multilingual safety alignment is still limited. After applying our MPO method---with English as the dominant language, given its well-established safety alignment---we observe a significant improvement in safety performance on X-Advbench across all 7 languages.

\end{document}